\documentclass[sn-mathphys-num]{sn-jnl}% Math and Physical Sciences Numbered Reference Style 
\usepackage{graphicx}%
\usepackage{multirow}%
\usepackage{amsmath,amssymb,amsfonts}%
\usepackage{amsthm}%
\usepackage{mathrsfs}%
\usepackage[title]{appendix}%
\usepackage{xcolor}%
\usepackage{textcomp}%
\usepackage{manyfoot}%
\usepackage{booktabs}%
\usepackage{algorithm}%
\usepackage{algorithmicx}%
\usepackage{algpseudocode}%
\usepackage{listings}%
\usepackage{comment}
\usepackage{subcaption}

\usepackage{siunitx}
\theoremstyle{thmstyleone}%
\theoremstyle{thmstyletwo}%

\theoremstyle{thmstylethree}%

\begin{document}

\title[Article Title]{Investigating the Impact of Communication-Induced Action Space on Exploration of Unknown Environments with Decentralized Multi-Agent Reinforcement Learning}

\author*[1]{\fnm{Gabriele} \sur{Calzolari}}\email{gabriele.calzolari@ltu.se}
\author[1]{\fnm{Vidya} \sur{Sumathy}}\email{vidya.sumathy@ltu.se} 
\author[1]{\fnm{Christoforos} \sur{Kanellakis}}\email{christoforos.kanellakis@ltu.se} 
\author[1]{\fnm{George} \sur{Nikolakopoulos}}\email{george.nikolakopoulos@ltu.se} 

\affil[1]{\orgdiv{Department of Computer Science, Electrical and Space Engineering}, \orgname{Luleå University of Technology}, \city{Luleå}, \country{Sweden}}


\abstract{This paper introduces a novel enhancement to the Decentralized Multi-Agent Reinforcement Learning (D-MARL) exploration by proposing communication-induced action space to improve the mapping efficiency of unknown environments using homogeneous agents. Efficient exploration of large environments relies heavily on inter-agent communication as real-world scenarios are often constrained by data transmission limits, such as signal latency and bandwidth. Our proposed method optimizes each agent's policy using the heterogeneous-agent proximal policy optimization algorithm, allowing agents to autonomously decide whether to communicate or to explore, that is whether to share the locally collected maps or continue the exploration. We propose and compare multiple novel reward functions that integrate inter-agent communication and exploration, enhance mapping efficiency and robustness, and minimize exploration overlap. This article presents a framework developed in ROS2 to evaluate and validate the investigated architecture. Specifically, four TurtleBot3 Burgers have been deployed in a Gazebo-designed environment filled with obstacles to evaluate the efficacy of the trained policies in mapping the exploration arena.}




\maketitle

\section{Introduction}\label{sec2}

Autonomous mobile robots are increasingly designated for the collaborative exploration of unknown environments, especially in inaccessible or hazardous environments preventing humans from mapping them. Space monitoring \cite{rl_swarm_robots, networked_multi_robot_exploration}, underwater surveillance \cite{Zhu2023MultiRobotAE_survey}, subterranean cave exploration \cite{tian2024iherointeractivehumanorientedexploration} and, notably, public safety scenarios and search and rescue missions \cite{drl_niroui} are some of the most relevant fields that can benefit from the deployment of autonomous robots. As for the last mentioned application, it can be highly advantageous for first responders and disaster management teams to quickly acquire maps of the disaster-affected areas while ensuring the safety of human lives. In such time-sensitive applications, minimizing the required exploration time, and ensuring the robustness and high degree of autonomy of the task execution are essential factors. 

\begin{figure}[t]
    \centering
    \includegraphics[width=0.6\linewidth]{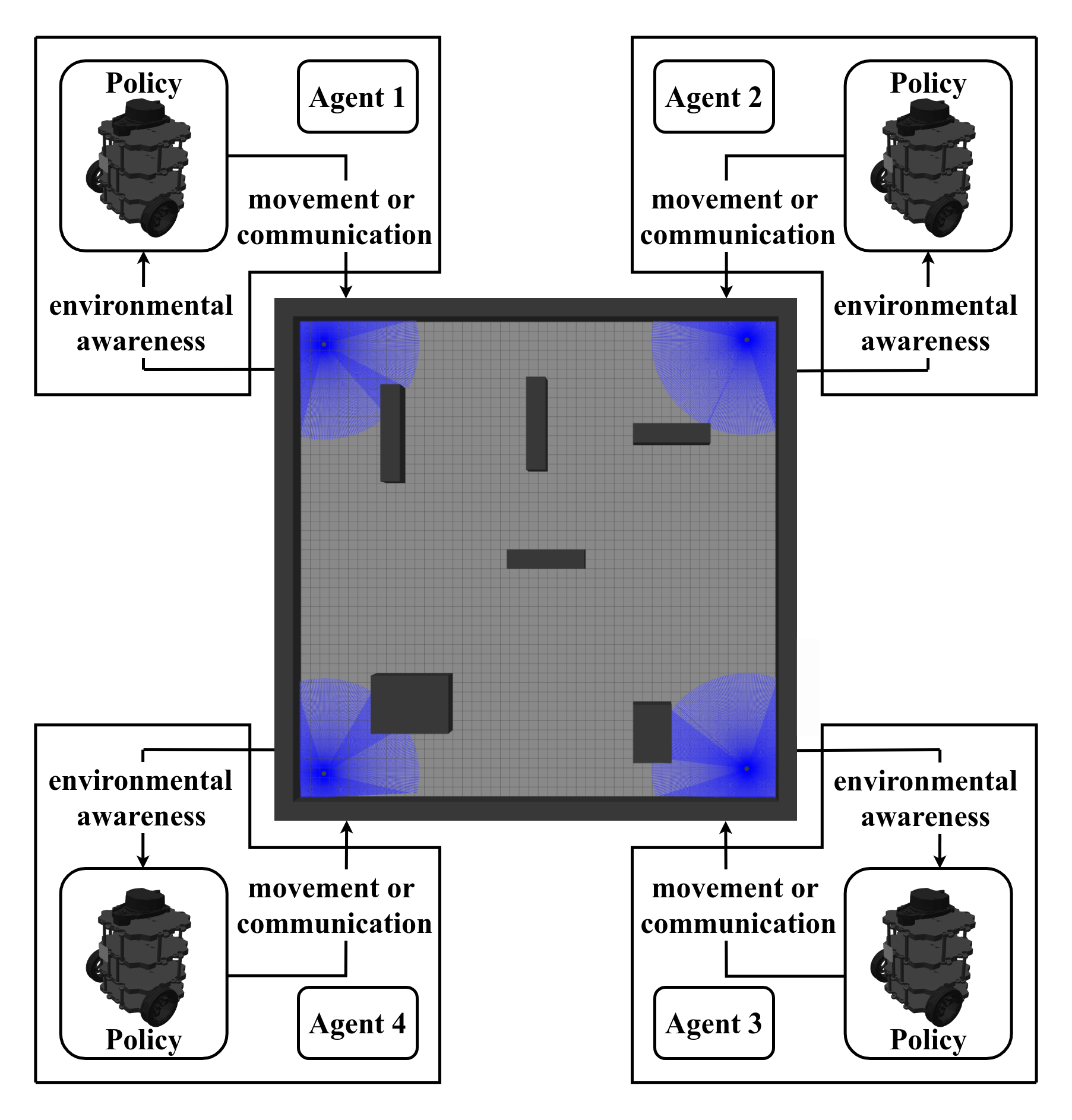}
    \caption{High-level overview of the exploration scheme where four TurtleBot3 Burgers navigate a Gazebo-designed environment with obstacles (dark gray). Each agent's policy decides the action to perform on the environment according to the morphology of the robot's neighborhood (blue) inferred through sensors}
    \label{fig:main_concept_image}
\end{figure}

In real-world scenarios, it should be assumed that the agents have very little or no prior knowledge about the exploration arena. Consequently, a suitable strategy should be defined such that the robots navigate to different locations to acquire information on the internal structure of the environment. Due to low efficiency, reliability, and complexity issues, multi-agent collaborative exploration is arousing more interest compared to single-agent exploration \cite{learning_exploration_deng, Zhu2023MultiRobotAE_survey}. However, this approach implies a detailed analysis of how inter-agent cooperation can be achieved. Some challenges concern decision-making for multi-agent navigation, control, localization, and inter-agent communication \cite{learning_exploration_hu}. Specifically, the constraints that affect data transmission can impair centralized collaborative architectures in real-world scenarios since these frameworks rely on a central node that is supposed to communicate with the operating agents. Indeed, the distance between the transmitter and receiver impacts signal attenuation, latency, and error rates, making communication between nearby agents typically the only sufficiently reliable option for safety-critical exploration applications. Furthermore, employing an event-based communication strategy helps to reduce unnecessary power consumption by data transmission devices. For these reasons, decentralized architectures are advantageous since agents can make autonomous decisions and collaborate through inter-agent communication, which enhances information sharing and improves the execution of exploration tasks. The scheme presented in Fig. \ref{fig:main_concept_image} is an example of an environment filled with obstacles and four Turtlebot3 Burger platforms to map the environment. In particular, each agent infers knowledge of the environment through its sensors and this information is elaborated by the robot's policy to enable some actions that can lead to movement in the environment or communication with other agents. 

Several strategies have been investigated to accomplish multi-agent exploration \cite{QuattriniLi2020_survey, Zhu2023MultiRobotAE_survey, survey_robotic_exploration_for_mapping}. Generally, these approaches may be classified into non-learning-based methods and learning-based ones. Particularly relevant methods belonging to the former class are frontier-based approaches due to their widespread usage \cite{frontier_based_exploration_yamauchi, frontier_based_exploration_greedy, frontier_based_exploration_cost, frontier_based_exploration_holz, frontier_based_batinovic, efficient_frontier_detection}. These algorithms reduce the number of map locations that need to be investigated by using the depth-first-search (DFS) algorithm \cite{frontier_based_exploration_yamauchi}, and cost or utility functions \cite{frontier_based_exploration_greedy, frontier_based_exploration_cost}. Some cooperative frontier-based approaches have been developed taking into account inter-agent communication to aid the exploration and optimize the mapping process \cite{cooperative_frontier_based_mahdoui}. However, the majority of these methods depend on a central node to allocate agents to different goals, operating under the assumption of effective communication among these entities. Some proposed multi-agent exploration strategies exploit information-based strategies to improve the mapping process \cite{mapping_exploration_bai, networked_multi_robot_exploration}. Recently, learning-based collaborative exploration strategies have drawn increasing attention due to their ability to deal with high-dimensional decision spaces. However, some learning-based features have been used to enhance frontier-based approaches, such as \cite{learning_exploration_burgard} which individuates frontier points and assigns them a cost according to the history of the exploration process.

In literature, many learning-based approaches for multi-agent exploration that use reinforcement learning to train the agents' policies have been investigated \cite{iqbal2020coordinated, rl_swarm_robots, learning_exploration_deng, learning_exploration_zhang}. Specifically, \cite{learning_exploration_hu} proposes an architecture to be deployed in a team of networked robots based on assigning different Voronoi partitions to the robots to avoid overlapping. Reinforcement learning is used here to allow the robot to reach the goals given by the high-level decision-making layer of the controller and to implement collision avoidance. \cite{drl_niroui} unifies frontier exploration with an A3C architecture to improve the exploration strategy obtained and deal with the high-dimensional robot states. Instead, \cite{learning_exploration_deng} proposes a frontier-based method that relies on a centralized training and decentralized architecture (CTDE), and the policies are trained using the multi-agent deep deterministic policy gradient (MADDPG) algorithm. Eventually, \cite{learning_exploration_zhang} models the environment as a topological graph and proposes the Hierarchical-Hops Graph Neural Networks (H2GNN), a cooperative decision-making framework that integrates information from the graph, and a variant of multi-agent proximal policy optimization (MAPPO) is used to learn the collaborative exploration strategy. However, most of these techniques assume continuous or periodic availability of inter-agent communication. \cite{commattn} proposes the Attention-based Communication neural network (CommAttn) to design a decentralized multi-agent architecture for the collaborative exploration of unknown environments by using explicit inter-agent communication. This network avoids useless data transmission between agents but does not consider limitations that affect communication. \cite{message_aware_graph_attention} investigates a decentralized message-aware graph attention network able to aggregate message features during communication while dealing with dynamic communication graphs during multi-agent exploration. \cite{cooperative_navigation_hu} proposes a cooperative approach for navigation with the exchange of information between agents in the same network and considering randomized dynamics parameters to make the policy adaptive to real-world scenarios.

Aiming at realistically modeling the inter-agent communication links, some decentralized strategies have been designed to consider limitations and constraints on data transmission that can happen in real-world scenarios \cite{survey_exploration_communication, effect_communication_restricted}. \cite{tolstaya2021multirobot} proposes a scalable Graph Neural Network (GNN) architecture that takes into account inter-agent proximity in dealing with information that can be shared between agents by changing the model's receptive field. Furthermore, the authors demonstrate that the proposed scheme can be trained using Reinforcement Learning, particularly the Proximal Policy Optimization (PPO) algorithm.

Acknowledging the existing literature, the work proposed in this research enhances the homogeneous agents' action space with communication. Therefore, each agent can choose to either share its occupancy grid map, which details the environment's structure, with other robots in the same communication network or decide which direction to move to continue exploring the arena. This is achieved by training the agents' policies using reward functions that account for inter-agent communication and efficient exploration. Furthermore, inter-agent transmission of local knowledge of the map follows a network-based approach between communicating agents. To the authors' knowledge, this is the first time such an approach has been proposed. The main contributions of this work include:

\begin{itemize}
    \item A novel decentralized collaborative multi-agent reinforcement learning (D-MARL) architecture with inter-agent communication integrated into the action space and a unique map-based observation space. 
    \item We propose and investigate different reward functions to study the efficacy of communication in the action space. 
    \item A framework developed in ROS2 and Gazebo for evaluating the proposed architecture with simulated TurtleBot3 Burgers.
\end{itemize}

\section{Preliminaries}

This section presents the general structure of the reinforcement learning problem that has been formulated to model the multi-agent exploration in \ref{subsection_reinforcement_learning_model}, the structure of the exploration arena which is represented as an occupancy grid in \ref{subsection_exploration_arena}, the maps that the autonomous agents use to represent the environment in \ref{subsection_agents_representation_of_the_environment} and the implementation details of the inter-agent communication strategy in \ref{subsection_communication_strategy}.

\begin{figure*}[t]
	%\centering
	\begin{subfigure}{0.23\linewidth}
		\includegraphics[width=\linewidth]{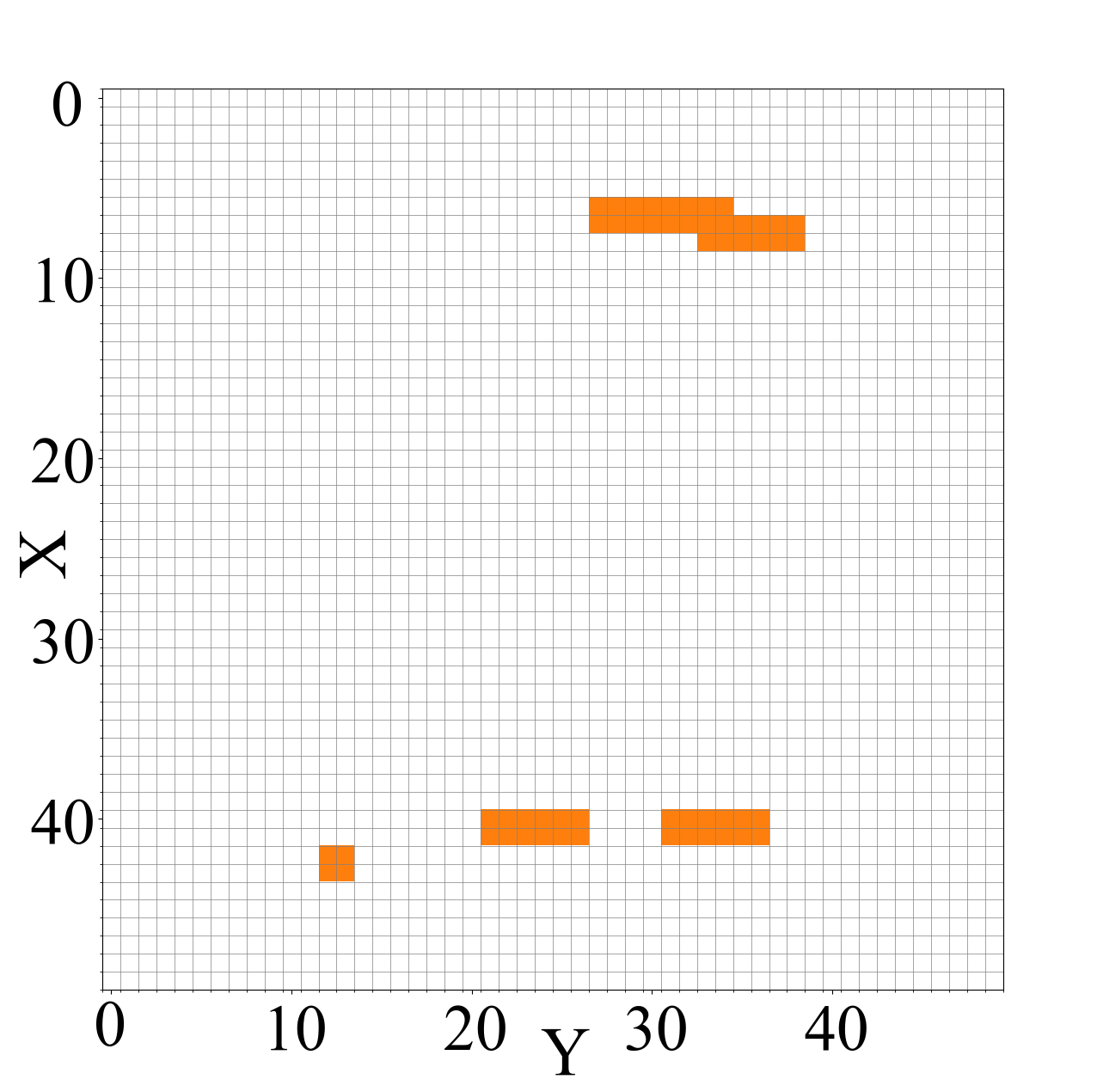}
		\caption{Arena with obstacles}
		\label{fig:arena_with_obstacles}
	\end{subfigure}%
        \hfill
	\begin{subfigure}{0.23\linewidth}
		\includegraphics[width=\linewidth]{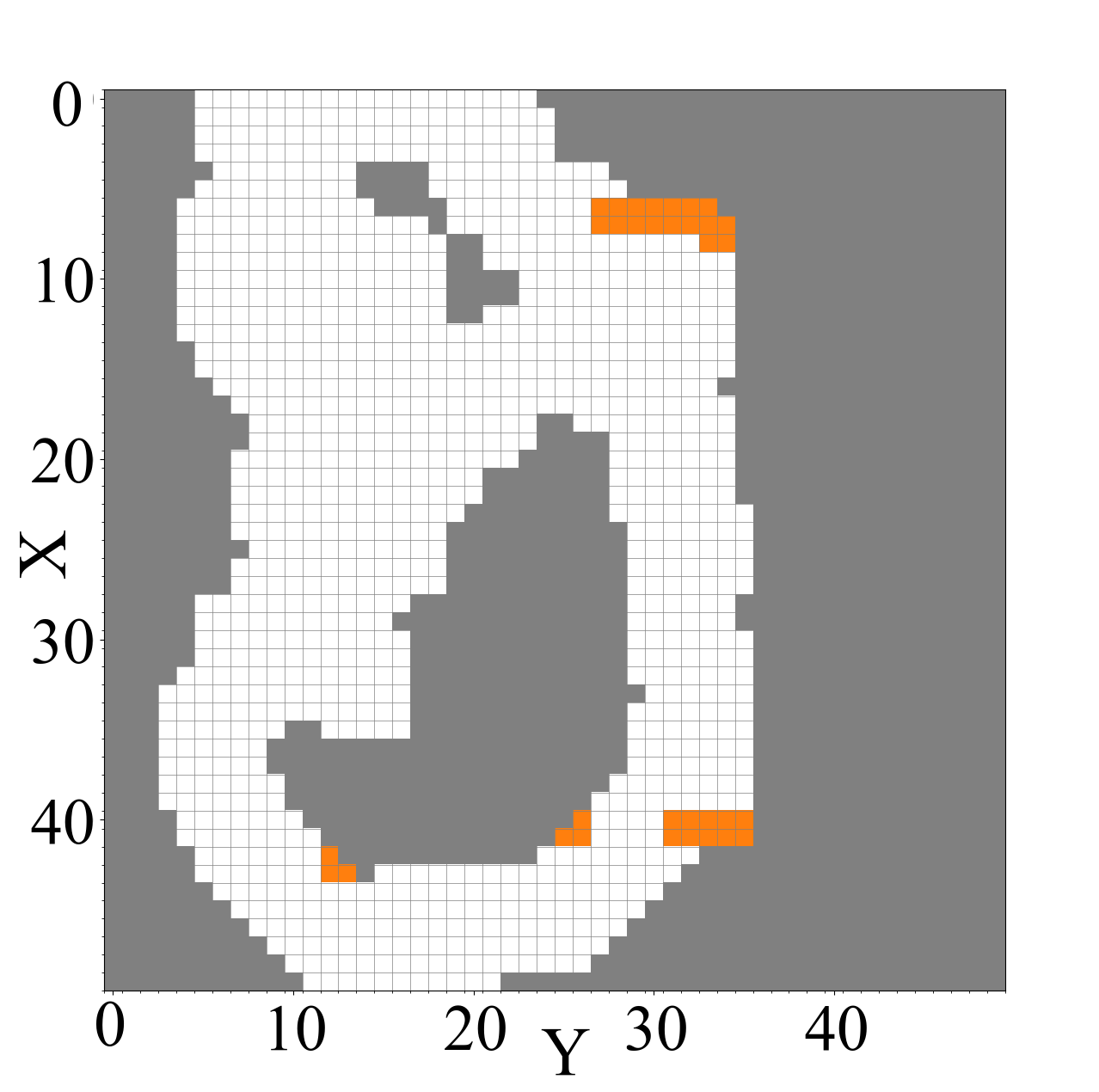}
		\caption{Agent-specific map}
		\label{fig:agent_local_map}
	\end{subfigure}%
        \hfill
        \begin{subfigure}{0.23\linewidth} 
            \includegraphics[width=\linewidth]{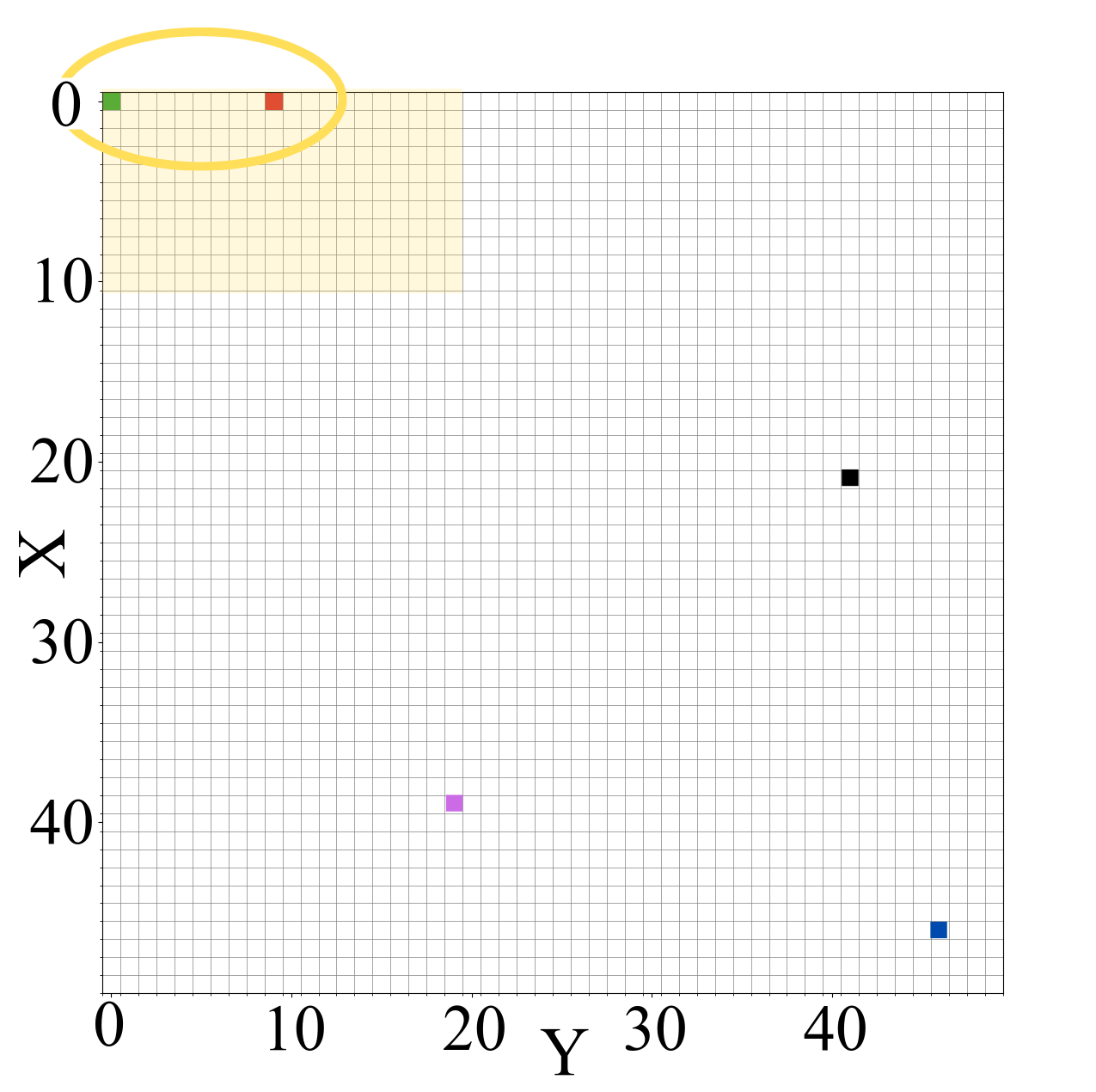}
            \caption{Transmission map}
            \label{fig:agent_transmission_map}
        \end{subfigure}%
        \hfill
        \begin{subfigure}{0.23\linewidth} 
            \includegraphics[width=\linewidth]{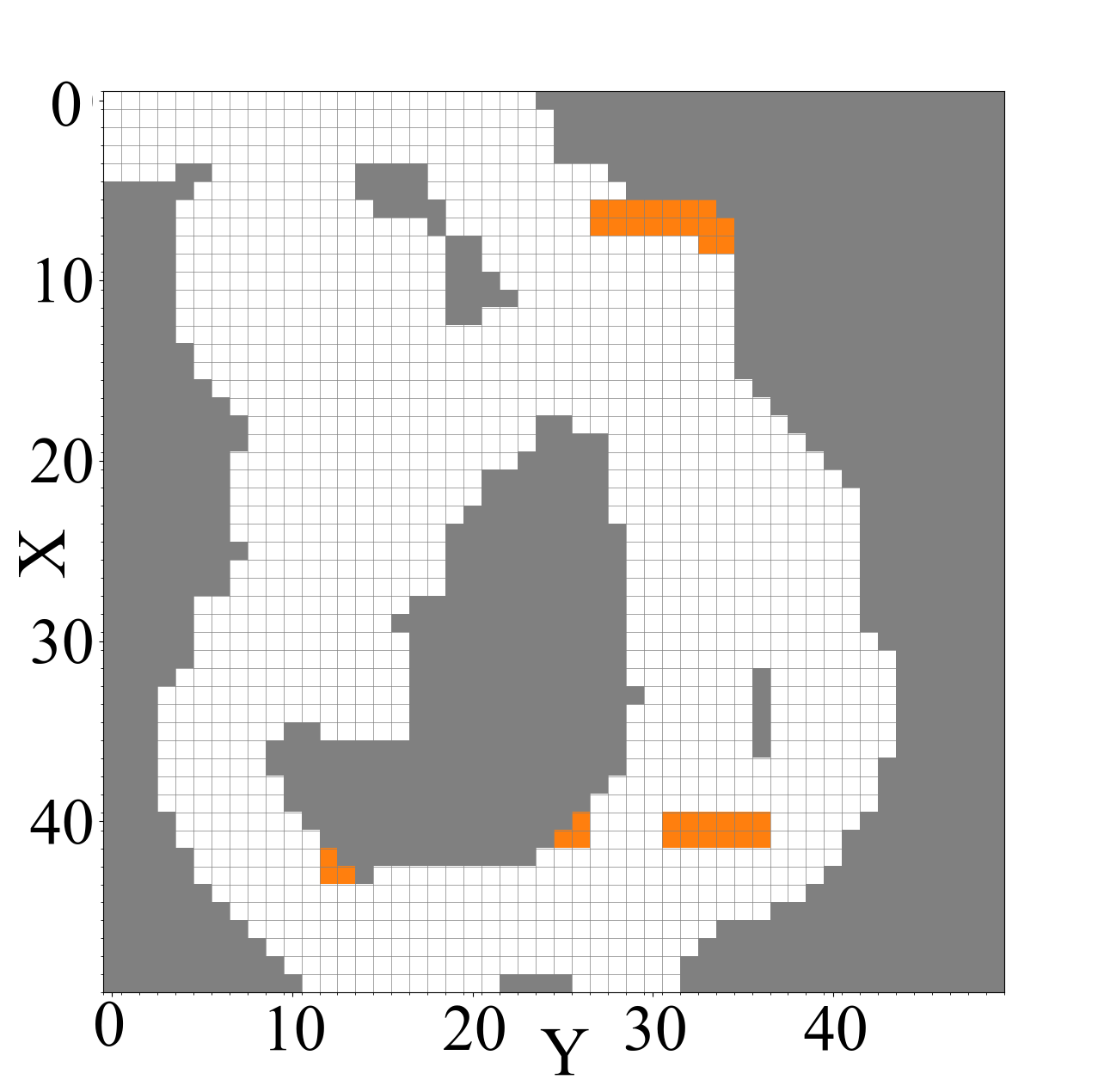}
            \caption{Collaborative map}
            \label{fig:agent_global_map}
        \end{subfigure}%
   
	\caption{The environment is represented as a 2D grid map, wherein the free, occupied, and undiscovered cells are denoted by white, orange, and gray cells, respectively. The agents are denoted by $\square$ markers with colors red, green, blue, black, and purple. Fig. \ref{fig:arena_with_obstacles} depicts a completely known exploration arena showing the free and occupied cells. Each robot has its agent-specific map as shown for the red agent in Fig. \ref{fig:agent_local_map}. While the communication network and communication covered area of the red agent are shown in Fig. \ref{fig:agent_transmission_map} by the yellow ellipse and shaded area. Finally, Fig. \ref{fig:agent_global_map} shows the red agent's collaborative map, obtained after blending the agent-specific maps of the communicating agents, namely the green and red ones}
	\label{fig:environment_representation}
\end{figure*}

\subsection{Reinforcement learning framework}
\label{subsection_reinforcement_learning_model}
We formulate the multi-agent exploration problem as a Partially Observable Stochastic Game (POSG) since the agents cannot directly sense the environment state and the actions chosen by the other agents. The POSG model is defined by the tuple $\textlangle \mathcal{I},\mathcal{S},\{\mu\},\{\mathcal{A}_k\},\{\mathcal{O}_k\},\{\textit{r}_k\},\mathcal{T}\textrangle$, where:
\begin{itemize}
    \item $\mathcal{I} = \{\mathit{i}_k, k \in \mathbb{N} \mathrel{|} 1 \leq k \leq n\}$ represents the finite set of the $n$ homogeneous agents interacting within the environment. At every time instant, each agent $i_k$ occupies a grid of the arena, described in \ref{subsection_exploration_arena}, and can detect the cells in a rectangular neighborhood with padding equal to $\textit{r}_d$. Moreover, each agent can directly communicate with the agents located within a squared region centered in the agent $i_k$ and with padding equal to the communication range $\textit{r}_c$.
    \item $\mathcal{S}$ represents the finite set of possible states, each of them represented by the occupancy map of the exploration arena, with grid states updated to reflect the discoveries made by all agents. Furthermore, each state keeps track of the agents' positions in the arena.
    \item $\{\mu\}$ represents the initial state distribution from which to draw the starting configuration of the agents in the arena. All the dispositions in which the agents are not arranged in occupied cells, i.e., in dangerous positions, are equally and uniformly possible.
    \item $\{\mathcal{A}_k\}$ and $\{\mathcal{O}_k\}$ represent respectively the finite set of actions and observations associated with agent $i_k$ and are described in \ref{subsection_action_space} and \ref{subsection_observation_space}.
    \item $\{\textit{r}_k\}$ is the reward function associated with the agent $i_k$ and a detailed description of the investigated ones is proposed in \ref{subsection_synthesized_reward_functions}.
    \item $\mathcal{T}$ denotes the probability that taking joint action $\textbf{a}_t$ in state $s_t$ results in a transition to a new state $s_{t+1}$ and joint observation $\textbf{o}_{t+1}$. In the setup proposed in this paper, the function $\mathcal{T}$ is deterministic. Indeed, the joint action and the current system state determine the next state and observations. 
\end{itemize}

The agent's policies have been designed as Convolutional Neural Networks, as discussed in \ref{paragraph_agents_policy}, and their parameters have been optimized through the Heterogeneous-Agent Proximal Policy Optimisation (HAPPO) algorithm \cite{heterogeneous_marl}.

\subsection{Exploration arena}
\label{subsection_exploration_arena}
In this work, the exploration arena is represented as a two-dimensional occupancy grid map $\mathcal{M}$ of size $\textit{n}$ $\times$ $\textit{n}$ with each cell of dimension $\textit{l}$ $\times$ $\textit{l}$. During exploration, every cell $\mathit{m}_{\mathit{i},\mathit{j}}$ is assigned a state value from $\mathcal{S} = \{0, 1, 2\}$, depending on whether the cell is without obstacles, occupied, or not yet explored, respectively. During exploration, all the free cells must be connected for the agents to map exhaustively the environment. Fig. \ref{fig:arena_with_obstacles} shows an example of the two-dimensional occupancy grid map for modeling the arrangement of the obstacles in the exploration arena.

\subsection{Agent's representation of the environment}
\label{subsection_agents_representation_of_the_environment}
In the proposed approach, to achieve the optimal course of action, each agent uses three maps: two maps of the environment, i.e., the agent-specific map and collaborative map, and the transmission map containing the agents' positions with which direct communication is possible. Fig. \ref{fig:environment_representation} illustrates the above-mentioned three maps of an agent and are defined as:
\begin{itemize}
    \item The \textit{Agent-Specific Map}, $\mathcal{M}_{i_k,as}$, is an occupancy grid with a global reference frame that contains only the information gathered from the agent's exploration.
    \item The \textit{Collaborative Map}, $\mathcal{M}_{i_k,co}$, collects the local observations contained in $\mathcal{M}_{i_k,as}$ and the information gathered by the other agents whenever a communication link is established.
    \item The \textit{Transmission map}, $\mathcal{M}_{i_k,tr}$, stores the positions of the agents that can be reached through direct communication in the given environment. 
\end{itemize}

\subsection{Communication strategy}
\label{subsection_communication_strategy}
Each agent $i_k$ can share its collaborative map $\mathcal{M}_{i_k,co}$ with other agents in the same communication network through two communication schemes: direct and indirect data transmission. Two agents can directly communicate with each other if and only if each agent is in the communication-covered area of the other agent. Moreover, a single agent can indirectly communicate with a distant agent by utilizing a chain of directly communicating agents.
When two or more agents communicate, they share all their collaborative maps and individually merge them into a single map. The newly generated map containing the discoveries of all the communicating agents is then used to update the agents' collaborative maps. Let a communication network be defined as $\mathcal{C}$, the merged maps are assigned to the agents' collaborative maps according to Eq. \eqref{eq_communication_formula}.

\begin{equation}
\label{eq_communication_formula}
\mathcal{M}_{i_k,co} = \bigcup_{i_j \in \mathcal{C}} \mathcal{M}_{i_j,co} \quad \forall \: i_k \in \mathcal{C}
\end{equation}

\begin{figure}[t]
    \centering
    \includegraphics[width=0.5\linewidth]{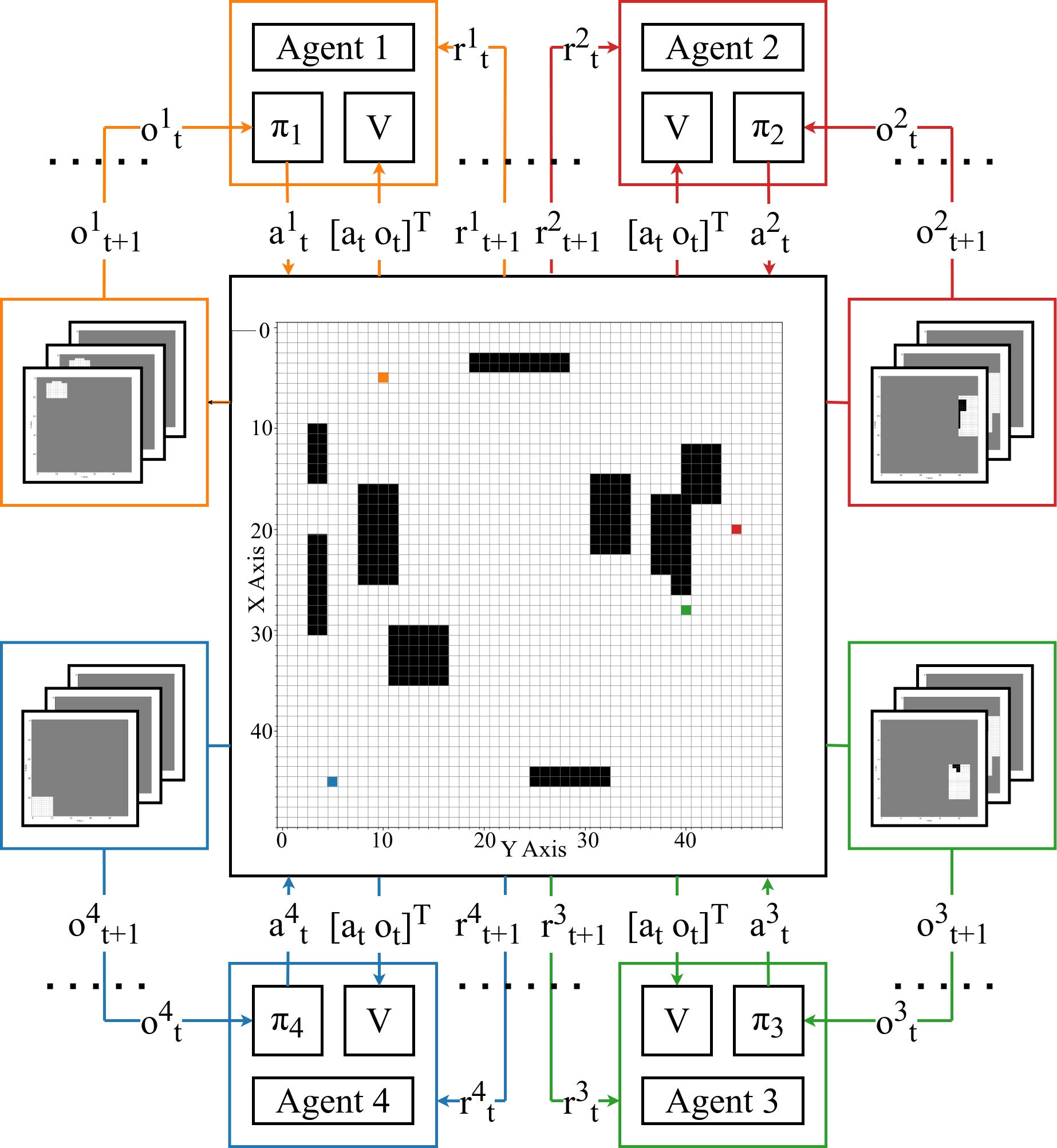}
    \caption{Proposed decentralized cooperative multi-agent reinforcement learning architecture for exploration. The occupancy grid depicts the environment with obstacles (black) and free cells (white), while the agents' positions are outlined by the colored cells. Furthermore, the components of each agent are shown, namely the agent-based policy ($\pi_k$) and the shared critic ($V$)}
    \label{fig:reinforcement_learning_scheme}
\end{figure}

\section{Methodology}
This section describes the proposed D-MARL exploration architecture, with its novel observation and action spaces. Fig. \ref{fig:reinforcement_learning_scheme} depicts the complete architecture of the proposed methodology with the four agents exploring an unknown environment represented by an occupancy grid.

\subsection{Action Space} 
\label{subsection_action_space}
One of the contributions of this work is the action space as defined in Eq. \eqref{eq:joint_action_space} and Eq. \eqref{eq_single_action}. In conventional approaches, the actions associated with a movement have 8 directions: up ($\uparrow$), up-right ($\nearrow$), right ($\rightarrow$), down-right ($\searrow$), down ($\downarrow$), down-left ($\swarrow$), left ($\leftarrow$) and up-left ($\nwarrow$). In addition to these movements, we propose two more actions that do not change the agent's position: an action called stay ($stay$), which allows the agents to remain stationary, and another called communicate ($comm$) which enables inter-agent communication according to the strategy explained in \ref{subsection_communication_strategy}. Equation \eqref{eq:joint_action_space} depicts the joint action at time $t \in \mathbb{N}$ and consists of a tuple containing the actions $a_t^{i_k}$ associated with each agent $i_k$. Each action $a_t^{i_k}$ belongs to a set of possible actions $\mathcal{A}_k$ as per Eq. \eqref{eq_single_action}.
    \begin{equation}
    \label{eq:joint_action_space}
        \textbf{a}_t = (a_t^{i_1}, ..., a_t^{i_n})
    \end{equation}
    \begin{equation}
    \label{eq_single_action}
    a_t^{i_k} \in \mathcal{A}_k = \{\uparrow, \nearrow, \rightarrow, \searrow, \downarrow, \swarrow, \leftarrow, \nwarrow, stay, comm\}
    \end{equation}

\begin{figure*}[t]
    \centering    
    \includegraphics[width=\linewidth]{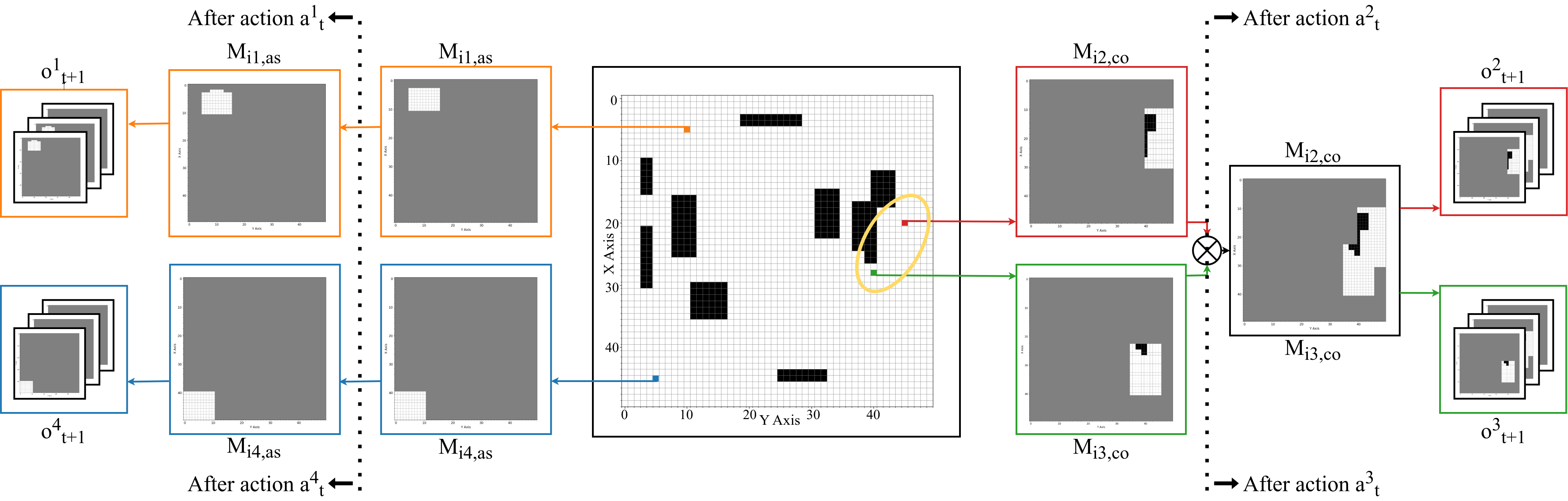}
    \caption{Generation of the agents' observations resulting from the execution of their actions within the environment. Specifically, the illustration shows agent $i_1$ (orange), agent $i_2$ (red), agent $i_3$ (green), and $i_4$ (blue) performing different actions and the updates affecting the policies' observations}
    \label{fig:generation_of_agents_observations}
\end{figure*}

\subsection{Observation space}
\label{subsection_observation_space}
Each agent's observation space consists of a 3D matrix that contains the agent-specific, collaborative, and transmission maps defined in \ref{subsection_agents_representation_of_the_environment}. Figure \ref{fig:generation_of_agents_observations} illustrates the process of generating the observations for all four agents interacting in the environment after having performed different actions. Specifically, the image shows agent $i_1$ (orange) performing one step upwards, while agent $i_2$ (red) and agent $i_3$ (green) communicating, and agent $i_4$ (blue) staying in the same position. The agent-specific map of $i_1$ shows that the known area is enlarged after the action is executed, while the agent-specific map of $i_4$ does not change since its position remains unchanged. Agents $i_2$ and $i_3$'s collaborative maps are instead merged since the two agents can communicate and the resulting map is assigned to both agents. The set of all three maps: agent-specific, collaborative, and transmission maps, are the observations given to the agents.

\begin{algorithm*}
\caption{Heterogeneous-Agent Proximal Policy Optimisation (HAPPO) algorithm \cite{heterogeneous_marl}}
\label{algorithm_happo}
\begin{algorithmic}
\Require number of agents $n \in \mathbb{N}$, number of episodes $n_e \in \mathbb{N}$, steps per episode $n_t \in \mathbb{N}$, batch size $n_b \in \mathbb{N}$, actor networks $\{\theta_{1}^{i_k} \mid i_k \in \mathcal{I}\}$, critic network $\{\phi_{1}\}$, replay buffer $\mathcal{D}$.
\For{$j = 1,\ldots ,n_e$}
\State $\mathcal{T} \gets $ generated trajectories using the joint policy $\boldsymbol{\pi_{\theta_j}} = (\pi^{1}_{\theta^{i_1}_j}, \ldots, \pi^{n}_{\theta^{i_n}_j})$ 
\State $\mathcal{D} \gets $ sampled transitions $\{(s_t, o^{i_k}_t, a^{i_k}_t, r_t, s_{t+1}, o^{i_k}_{t+1}), {i_k}\in \mathcal{I}, t \in n_t\}$ from $\mathcal{T}$
\State $\mathcal{M} \gets \text{RandomSample}(\mathcal{D})$
\State The advantage function $\hat{A}(s, \boldsymbol{a})$ is computed using the critic network with Generalized Advantage Estimator (GAE).
\State $i_{1:n} \gets \text{permutation}(i_{1:n})$
\State $M^{i_1}(s, \boldsymbol{a}) \gets \hat{A}(s, \boldsymbol{a})$
\For{$i_k \in i_{1:n}$}
    \State $w_t^{i_k} \gets \frac{\pi^{i_k}_{\theta^{i_k}} (a_t^{i_k} | o_t^{i_k})}{\pi^{i_k}_{\theta^{i_k}_j} (a_t^{i_k} | o^{i_k}_t)}$
    \State $\theta_{j+1}^{i_k} \gets \operatorname*{argmax}_\theta \frac{1}{n_b
    n_t} \sum_{b=1}^{n_b} \sum_{t=0}^{n_t} \min \Biggl(w_t^{i_k} M^{i_{1:k}}(s_t, \boldsymbol{a}_t), \textit{clip} (w_t^{i_k} , 1 \pm \epsilon) M^{i_{1:k}}(s_t, \boldsymbol{a}_t) \Biggr)$ 
    \State $M^{i_{1:k+1}}(s, \boldsymbol{a}) \gets \frac{\pi^{i_k}_{\theta^{i_k}_{j+1}}(a^{i_k}|o^{i_k})}{\pi^{i_k}_{\theta^{i_k}_j}(a^{i_k}|o^{i_k})} M^{i_{1:k}}(s, \boldsymbol{a})$ 
\EndFor
\State $\phi_{j+1} \gets \operatorname*{argmin}_\phi \frac{1}{n_b n_t} \sum_{b=1}^{n_b} \sum_{t=0}^{n_t} \Biggl(V_\phi(s_t) - \hat{R}_t \Biggr)^2$ 
\EndFor

\end{algorithmic}
\end{algorithm*}

\subsection{Multi-Agent Reinforcement Learning algorithm}
\label{subsection_multi_agent_reinforcement_learning_algorithm}

Each agent is controlled by an independent policy whose parameters are optimized through the Heterogeneous-Agent Proximal Policy Optimisation (HAPPO) algorithm. The implementation of this technique has been presented in \cite{heterogeneous_marl}. Specifically, the policy optimization algorithm HAPPO is based on a multi-agent sequential update method. In this algorithm, there are two essential elements for each agent: a shared critic, $V(\phi)$, and a non-shared policy, $\pi^k(\theta^{i_k})$. The algorithm first generates some trajectories for the system and gathers samples. Consecutively, HAPPO updates each policy sequentially for $n_e \in \mathbb{N}$ episodes, such that the next agent’s advantage is iterated by the current sampling importance and the former advantage. Algorithm \ref{algorithm_happo} provides a detailed description of how HAPPO recurrently improves the policies and the critic networks. The values assigned to the hyper-parameters used in Algorithm \ref{algorithm_happo} are given in Table \ref{tab:training_params}. Moreover, key parameters and variables for the HAPPO algorithm include $\epsilon$ as the clipping threshold, $s_t$ and $\textbf{a}_t$ as the current global state and joint action respectively, $V_\phi(s_t)$ for the critic network's value function prediction, and $\hat{R}_t$ for the rewards-to-go. The architecture of the neural networks implementing the agents' policies and critic functions is discussed in detail in the next paragraphs. 

\begin{figure}[t]
    \centering
    \begin{subfigure}[b]{0.6\linewidth}
        \centering
        \includegraphics[width=\linewidth]{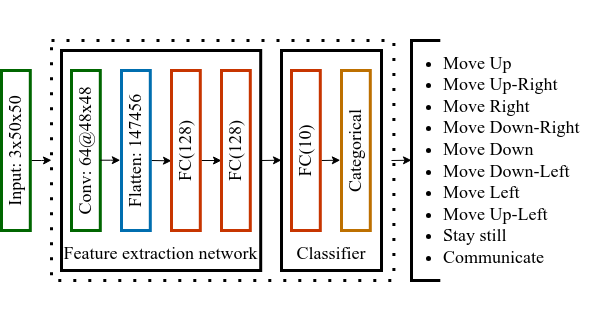}
        \caption{The architecture of the agents' policy network. The classifier selects an action from the ten options in the action space shown on the right}
        \label{fig:agent_policy_network}
    \end{subfigure}
    
    \begin{subfigure}[b]{0.6\linewidth}
        \centering
        \includegraphics[width=0.7\linewidth]{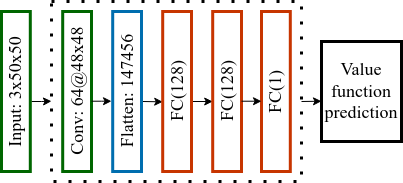}
        \caption{Agents' critic network architecture}
        \label{fig:agent_critic_network}
    \end{subfigure}
    
    \caption{The diagrams depict the architecture of the neural networks used as the agents' policies (\ref{fig:agent_policy_network}) and the shared critics (\ref{fig:agent_critic_network}). Each illustration details the input size, the core layers composing the networks, and the resulting output data. Layers of the same type are highlighted with the same colors: green for convolutional layers (Conv), blue for flatten layers, red for fully-connected layers (FC), and orange for categorical distributions. Furthermore, each component specifies the output size of its respective layer}
    \label{fig:network_architectures}
\end{figure}

\paragraph{Agent's policy} 
\label{paragraph_agents_policy} Each agent utilizes its trained policy to choose an action based on the observations received from the environment in the previous step. Fig. \ref{fig:agent_policy_network} illustrates the neural network architecture that models the agents' policies. As the picture shows, it consists of two sub-modules: a feature extraction network based on a convolutional layer, and a classifier implemented by using a multilayer perceptron (MLP). The former network takes in the three-layer observation of the environment described in \ref{subsection_observation_space} as input and applies a convolutional layer. Then, the obtained flattened feature image is processed by two fully connected layers to extract meaningful features from the agent's maps so that the classifier can choose an appropriate action from the discrete set $\mathcal{A}_k$ defined in Eq. \eqref{eq_single_action}. The convolutional layer is a fundamental component of the policy network and it works by applying the set of operations described in Eq. \eqref{eq:convolution} with hyper-parameters expressed in Table \ref{tab:hyperparameters_convolutional_layer}.

\begin{equation}
\label{eq:convolution}
Y = f\left( \sum_{i=1}^{C_{in}} \mathbf{W}_i \star \mathbf{X} + \mathbf{b} \right)
\end{equation}

where $\mathbf{X}$ is the input feature map with dimensions $3 \times 50 \times 50$ assuming that the parameters used to model the environment are those reported in Table \ref{tab:training_params}, $C_{in}$ is the number of channels in the input, $\mathbf{W}_i$ denote the trainable filters (kernels) associated with each channel,  $\star$ denotes the convolution operation, $\mathbf{b}$ stands for the bias term, $f$ is the rectified linear unit (ReLU) activation function, and $Y$ represents the feature map with dimensions $64 \times 48 \times 48$, as per the convolution hyperparameters in Tab. \ref{tab:hyperparameters_convolutional_layer}.

\begin{table*}[]
    \centering
    \caption{Hyper-parameters of the convolutional layer}
    \label{tab:hyperparameters_convolutional_layer}
    \begin{tabular}{lcccc}
        \toprule
        Parameter & Input channels ($C_{in}$)& Output channels & Kernel size & Stride\\
        \midrule
        Value & 3 & 64 & (3, 3) & (1, 1)\\
        \bottomrule
    \end{tabular}
\end{table*}

The output of the convolutional layer is then flattened into a one-dimensional tensor to be processed by two fully connected layers. Each layer produces 128 output features and is followed by the ReLU activation function. After the feature extraction module, the implemented classifier, which has been designed as a multilayer perceptron, allows selecting an action from the action space according to the features extracted by the previous layers. Specifically, the dense layer in this network outputs 10 values, as the cardinality of the action space $|\mathcal{A}_k|$, and is followed by a categorical distribution. Furthermore, the policy implementation enables the masking of specific actions during inference, given that the unavailable actions are adequately specified.

\paragraph{Critic Network}
Each agent is associated with a critic network used only during the training of the agents' policies. In contrast with the policy network, the critic needs to output just one value which is the value function prediction as Fig. \ref{fig:agent_critic_network} shows.

\subsection{Reward and Penalization terminology}
\label{paragraph_reward_function}
Intending to maximize non-repetitive exploration of the environment and prioritize network-based communication between the agents, we propose and compare different agent-based reward functions. Each reward has one or more of the terms defined below and is referred to each time step in the environment:
\begin{itemize}
    \item The exploration reward $r_{k,exp}$ is designed to persuade the agents to explore new grids compared to their collaborative map. This term is described by the following expression:
    \begin{equation}
        r_{k, exp} = \frac{\Delta \mathcal{M}_{i_k,co}}{\textit{e}_{max}} \in [0,1]
    \end{equation}
    where $\Delta \mathcal{M}_{i_k,co}$ denotes the number of grid cells that have been explored by the agent $i_k$ in the given time step, and that increases its knowledge about the environment. The parameter $\textit{e}_{max}$ represents the maximum increase possible in the knowledge of the environment due to a unitary step. This term depends on the characteristics of the agent and the environment and can be computed according to the following formula:
    \begin{equation}
        \textit{e}_{max} = 4 \frac{\textit{r}_d}{l} + 1
    \end{equation}
    where $l$ and $r_d$ are the sizes of the cells in the exploration arena and the agent's detection range, respectively, as illustrated in \ref{subsection_reinforcement_learning_model}.
    \item The communication reward $r_{k,com}$ is the term in the reward function that should encourage the agents to share the knowledge they have gathered during the exploration phase. This term is only available in the reward function when the agent $i_k$ communicates otherwise, its value is $0$ through the parameter $\textit{w}_{i_k}$. If the agent communicates, the reward is proportional to the increase of the knowledge that happens due to the communication with the agents in the same network ($c_{i_k,co}$) divided by the number of unknown grids in the map, i.e., $n_{unknown}$. Depending on the study case illustrated in \ref{subsection_synthesized_reward_functions}, the communication reward can be multiplied by a coefficient $p_k$ to prioritize this reward over the exploration one. The following formula describes the communication term in the reward function:
    \begin{equation}
        r_{k,com} = \textit{w}_{i_k} \frac{c_{i_k,co}}{n_{unknown}}
    \end{equation}

    \item The stay-still penalization $r_{k,rep}$ discourages the agent from staying stationary, except if the communication action has been enabled. If the agent $i_k$ remains still, it receives a reward of $r^*_{rep}$. 
    \item A penalization $r_{k,bou}$ if the agent $i_k$ is close to the grid map boundaries. If it occupies a cell near the boundary, a reward of $r^*_{bou}$ is assigned. 
    \item A penalization $r_{k,col}$ if the agent $i_k$ goes into an obstacle, another agent, or out of the map boundaries which is equal to $r^*_{col}$.
    \item A penalization $r_{k,ndi}$ that diminishes by $r^*_{ndi}$ as $r_{k, exp}$ is lower than $r^*_{exp}$. This term should reinforce the importance of actions that extend the knowledge about the environment. 
\end{itemize}
\subsection{Synthesized Reward Functions}
\label{subsection_synthesized_reward_functions}
To analyze the effectiveness of incorporating communication in the action space, different reward functions are formulated based on the reward and penalization terms described in \ref{paragraph_reward_function}. 
 Each of the proposed reward functions is discussed as a different study case. In every scenario, if the agent selects an action that implies the collision of the agent $i_k$ into an obstacle, another agent, or an arena boundary then the reward would be $r_{k} = r_{k,col}$.
\begin{itemize}
\item \textit{Study Case 1}: A baseline reward function considering only exploration and without any reward for inter-agent communication is designed. The proposed reward function is formulated as:
\begin{equation}
r_{k} = r_{k,exp} + r_{k,rep} + r_{k,bou} + r_{k,ndi} 
\end{equation}
\item \textit{Study Case 2}: In this case, the designed reward function is analogous to the one in \textit{Study Case 1}, but now the communication term is added. As mentioned in \ref{paragraph_reward_function}, this contribution to the reward prioritizes communication as it intends to expand the agents' knowledge avoiding useless transmissions of data. The following formula describes how the reward function is structured.
\begin{equation}
r_{k} =  r_{k,exp} + r_{k,com} + r_{k,rep} + r_{k,bou} + r_{k,ndi}
\end{equation}
\item \textit{Study Case 3}: A modified reward function is formulated in this case by adding a parameter $p_k$ multiplied by the communication reward to the reward function used in \textit{Study Case 2}. This is to take into account the number of cells explored in the previous step. The new coefficient is formulated in the following way:
\begin{equation}
    p_k = \frac{c_{i_k,co}(previous \: step)°}{\textit{e}_{max}} + 0.6
\end{equation}
Thus, the reward function is written as:
\begin{equation}
r_{k} =  r_{k,exp} + p_k r_{k,com} + r_{k,rep}  + r_{k,bou} + r_{k,ndi}
\end{equation}
\item \textit{Study Case 4}: In this reward function, the coefficient $p_k$ mentioned in \textit{Study Case 3} is changed with the average of the number of grids explored by the agent since the last communication with each agent participating in the same communication network. The new definition of the parameter $p_k$ is shown in Eq. \eqref{equation_rho_case_4}. 
\begin{equation}
\label{equation_rho_case_4}
    p_k = \frac{1}{N-1} \sum_{i_j \in \mathcal{I}, i_j \neq i_k} \frac{\textit{w}_{i_j}\textit{q}_{i_j}\textit{d}_{i_k,i_j}}{\textit{n}^2} + 0.8
\end{equation}
In this formulation, the variable $\textit{q}_{i_j}$ takes the value $1$ if the agent $i_j$ participates in the same communication network as agent $i_k$. The term $\textit{d}_{i_k,i_j}$ stands for the number of discoveries performed by agent $i_k$ since the last communication with agent $i_j$. Moreover, in this version, the boundary and negative exploration terms are not present, and the denominator of the communication reward $r_{k,com}$ is $\textit{e}_{max}$.
\begin{equation}
r_{k} = r_{k,exp} + p_k r_{k,com} + r_{k,rep} 
\end{equation}
\end{itemize}

\section{RL Policy Training in Gymnasium}
%This section illustrates how the agents' policies have been trained, the evaluation metrics, and compares the investigated reward functions to understand the effect of adding inter-agent communication in the agents' action spaces on the efficacy of the mapping process. Specifically, the proposed scheme has been evaluated using the four trained agents in randomly generated environments similar to the training ones with the characteristics mentioned in section \ref{subsection_exploration_arena}.

%\subsection{Training of the agents' policies}

\begin{figure}[h]
  \centering
  \includegraphics[width=0.5\linewidth]{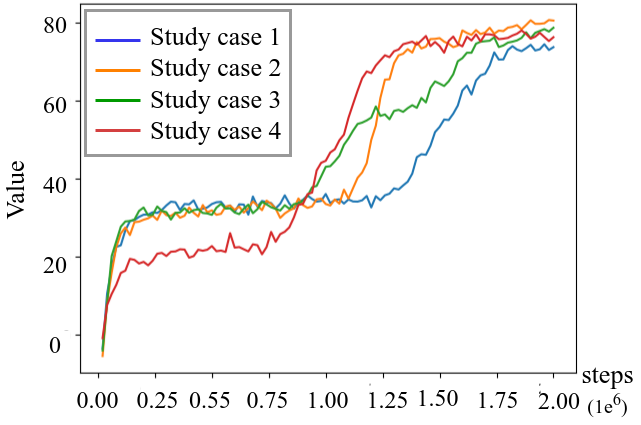} 
  \caption{Train episode reward curves for each study case}
  \label{fig:train_episode_reward}
\end{figure}

\label{subsection_training_setting}
The agents' policies are trained in a Ubuntu 20.04.6 LTS Computer with 32 13th Gen Intel Core i9-13900K and an NVIDIA GeForce RTX 4090. In particular, the training of the policies has been done using samples from environments with different obstacles and starting positions for the agents. Overall, the environment has been modeled using the PettingZoo library and the Parallel API \cite{pettingzoo}.  The learning rate has been set for the actor and critic networks at $5 \times 10^{-4}$ and the number of epochs for update is set to 5.  Table \ref{tab:training_params} contains the values associated with the most important parameters for the environment, the reward function, and the training process. Fig. \ref{fig:train_episode_reward} shows that the train episode reward curves obtained during the training of the policies developed according to the four study cases have similar behaviors. However, study case 4 achieves convergence of the episode reward more swiftly, while the worst one is represented by study case 1.

\begin{table}[h]
    \centering
    \caption{Hyper-parameters used for the environment, reward functions, and the training process}
    \label{tab:training_params}
    \begin{tabular}{lccccc}
        \toprule
        \multicolumn{6}{c}{\textbf{Environment Parameters}} \\
        \midrule
        Name & $n$ & $l$ & $r_d$ & $r_c$ \\
        \midrule
        Value & 50 & 0.5 & 1 & 5 \\
        \toprule
        \multicolumn{6}{c}{\textbf{Reward Parameters}} \\
        \midrule
        Name & $r^*_{\text{rep}}$ & $r^*_{\text{bou}}$ & $r^*_{\text{col}}$ & $r^*_{\text{ndi}}$ & $r^*_{\text{exp}}$ \\
        \midrule
        Value & -1 & -1 & -10 & 0.2 & 0.3 \\
        \toprule
        \multicolumn{6}{c}{\textbf{Training Parameters}} \\
        \midrule
        Name & $n_a$ & $n_e$ & $n_s$ & $n_b$ & $\epsilon$ \\
        \midrule
        Value & 4 & 10,000 & 200 & 1 & 0.2 \\
        \bottomrule
    \end{tabular}
\end{table}
\begin{table*}[t]
    \centering
    \caption{Metrics collected from the simulations in Gymnasium of the policies trained according to the investigated study cases on 100 environments belonging to the testing environments}
    \label{table_evaluation_cases_metrics}
    \begin{tabular}{l>{\centering\arraybackslash}p{1cm}>{\centering\arraybackslash}p{1cm}>{\centering\arraybackslash}p{1cm}>{\centering\arraybackslash}p{1cm}>{\centering\arraybackslash}p{1cm}>{\centering\arraybackslash}p{1cm}>{\centering\arraybackslash}p{1.7cm}}
        \toprule
         & \multicolumn{2}{c}{$n_{steps}$} & \multicolumn{2}{c}{$\mathcal{J}$} & \multicolumn{2}{c}{$D_{shared}$} & $robustness$ \\
        \cmidrule(lr){2-3} \cmidrule(lr){4-5} \cmidrule(lr){6-7} \cmidrule(lr){8-8}
         & Mean & Std & Mean & Std & Mean & Std & Value \\
        \midrule
        Case 1 & 557.00 & 169.07 & 0.28 & 0.07 & 475.50 & 132.18 & 0.60\\
        Case 2 & 487.04 & 233.63 & 0.25 & 0.05 & 489.70 & 127.46 & 0.64\\
        Case 3 & 511.00 & 168.10 & 0.25 & 0.05 & 0.00 & 0.00 & 0.79\\
        Case 4 & 454.33 & 184.85 & 0.25 & 0.05 & 574.43 & 116.48 & 0.86\\
        \bottomrule
    \end{tabular}
\end{table*}

\section{Simulation Results}
This section outlines the evaluation metrics and simulation results, assessing the effectiveness of the proposed reward functions. It aims to understand the impact of incorporating inter-agent communication into the agents' action space for mapping tasks. Specifically, the investigated scheme has been evaluated using four trained agents in randomly generated environments that resemble the training environments described in Section \ref{subsection_exploration_arena}. The analyzed methodology was simulated and evaluated in two distinct simulation platforms: Gymnasium and Gazebo, as presented below.
\subsection{Metrics} 
\label{subsection_metrics}
\cite{xu2022explorebench} has investigated useful criteria to evaluate reinforcement learning-based exploration. In this paper, the following metrics have been used to evaluate and compare the different reward functions mentioned in \ref{subsection_synthesized_reward_functions}:
\begin{itemize}
\item The total number of steps $n_{steps}$ required by the agents to explore the unknown environment. In these experiments, it has been assumed that an acceptable exploration terminates when $\textit{p} = 90\%$ of the grids without obstacles have been explored. 
\item The efficacy of the collaboration scheme between the agents is estimated from the overlap between the agent-specific maps gathered by each exploring agent. This parameter is quantified when the exploration process is concluded and is based on the Jaccard similarity coefficient \cite{costa2021generalizations}. Specifically, this statistical index is estimated as the average of the Jaccard indices between all pair combinations of the agent-specific maps, as below,
\begin{equation}
    \mathcal{J} = 2 \frac{\sum_{i_k \in \mathcal{I}} \sum_{i_j \in \mathcal{I}, \ j > k} J(\mathcal{M}_{i_k,as}, \mathcal{M}_{i_j,as})}{|\mathcal{I}| \times (|\mathcal{I}| - 1)}
\end{equation}
\item The effectiveness of communication can be represented by the quantity of information shared among the agents communicating in the same network. This metric is referred to as $D_{shared}$.
\item The robustness of the trained policies when dealing with different environments. Specifically, this metric is computed considering the ratio of the successfully explored arena over the total number of testing environments. 
\end{itemize}
\subsection{Simulation Results in Gymnasium}
\label{sbs:simulation_results_in_gymnasium}
\begin{figure}[h]
        \centering
        \includegraphics[width=0.5\linewidth]{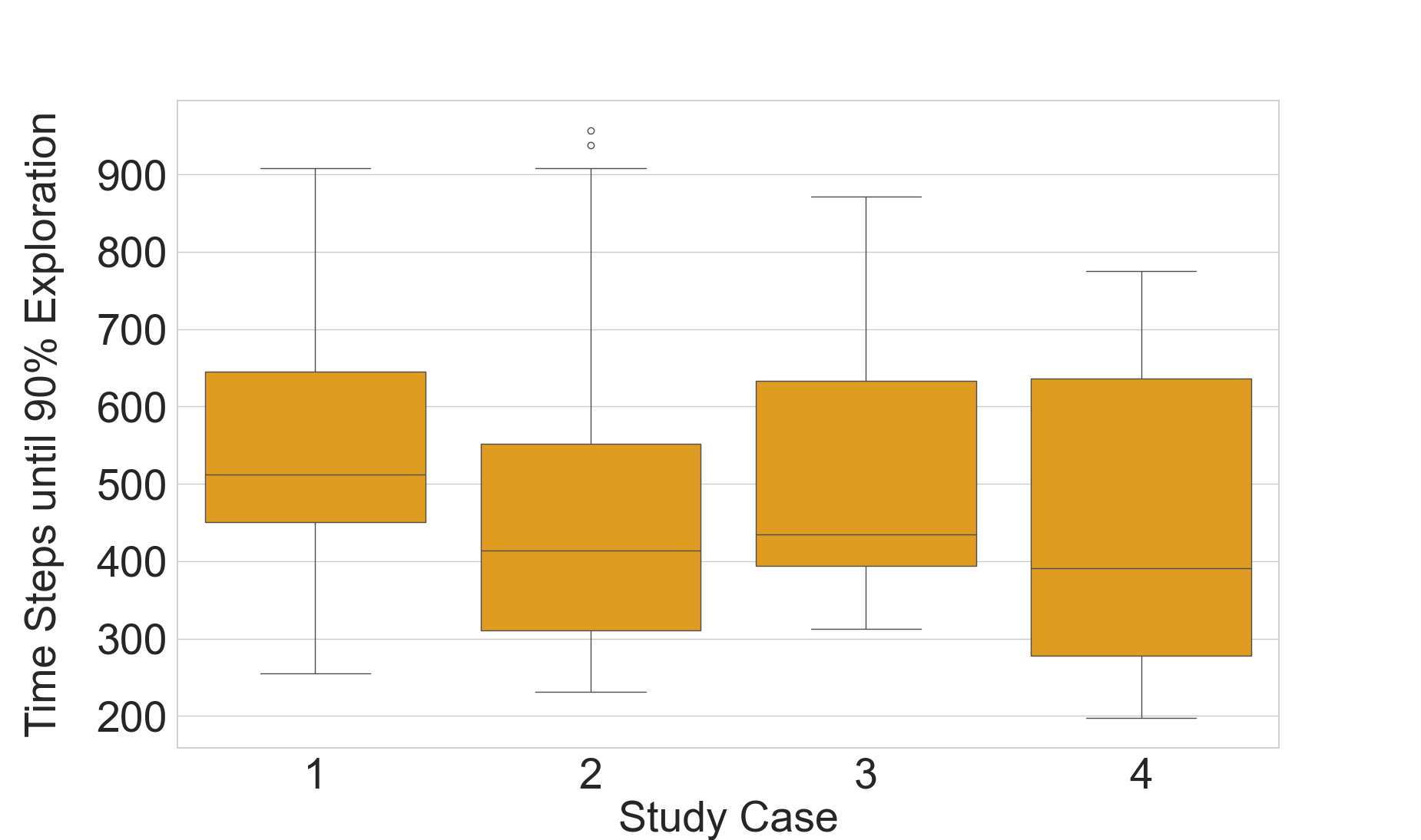} 
        \caption{Statistical distribution of the number of steps required to explore the testing environments according to the four study cases}
        \label{fig:n_steps_simulations}
\end{figure}

\begin{comment}
    Fig. \ref{fig:train_episode_reward} shows the episode rewards obtained during the training of the agents' policies according to the proposed study cases. It can be stated that all the curves follow a similar pattern since they are characterized by an increasing trend with a central plateau. Nonetheless, the figure shows that study cases 2 and 4 converge more swiftly than the others. Specifically, case 1 appears to be the slowest one in reaching the convergence. Furthermore, it can be observed that all the cases do not incur unstable behaviors and that eventually, they converge to solutions that achieve similar rewards.
\end{comment}
All four study cases have been evaluated in 100 different environments with obstacles at different locations. Fig. \ref{fig:n_steps_simulations} illustrates the distribution of the number of steps required by each study case to explore the testing environments. The results show that the policies trained with communication-based reward functions achieved exploration with a lower number of steps compared to study case 1. Also, study cases 2 and 4 succeeded in lowering the median of the number of steps by more than 100 steps compared to the case without communication. However, the interquartile range is larger since the collected data are more sparse. Furthermore, the results show that the median in the number of steps is significantly low for case 4, outperforming all the other study cases. 

Table \ref{table_evaluation_cases_metrics} shows the values of the metrics explained in \ref{subsection_metrics} and obtained from the simulations of the trained policies in the environments used to evaluate the different study cases. The results indicate that the policies developed through study case 4 exhibit the best robustness, achieving over 86\%. This suggests that these policies can successfully explore more environments compared to those from other study cases. Specifically, study case 1 has the least robustness as it could successfully explore only $60\%$ of the testing exploration arenas. 
\begin{figure*}[h]
  \centering
  \includegraphics[width=\textwidth]{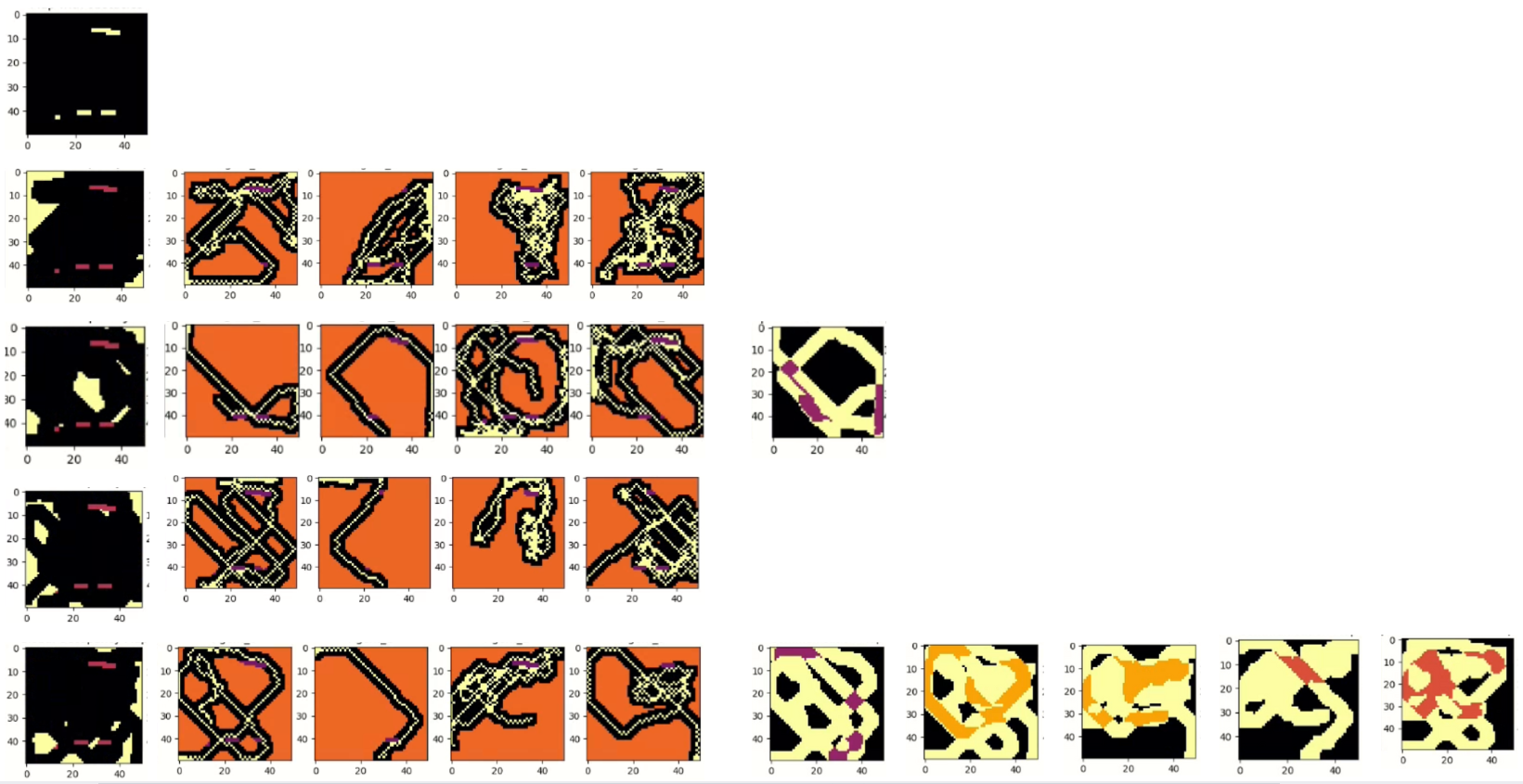} 
  \caption{Collection of exploration results performed by the study cases in an unknown environment with obstacles. The topmost image illustrates the free cells (black) and obstacles (yellow) that are in the environments. Each row contains the samples collected from the complete exploration done by the policies developed in each study case. In each series, the first image depicts the occupancy map of the environment after each exploration. In particular, they show the free cells (black), the obstacles (purple), and the unknown grids (yellow). Then, the trajectories of agents 1 to 4 are shown from left to right, respectively. These pictures show the cells covered by the trajectories (yellow), the direct discoveries of the agents (black), and the unknown cells (orange). The last images, whenever displayed, show the map that is shared between communicating agents in the same networks. The yellow areas depict non-overlapping areas between the agents, while the purple, yellow, and orange ones are the sections that overlap between the agents that participate in the communication scheme}
  \label{fig:case_data}
\end{figure*}

Considering the other metrics mentioned in \ref{subsection_metrics} and reported in Table \ref{table_evaluation_cases_metrics}, the averages of the Jaccard similarity coefficients representing the overlapping between the agent-specific maps gathered by the robotic platforms show that both cases 2, 3, and 4 obtain similar performances and improve the reduction in overlapping compared to study case 1. This distinctly proves the efficacy of adding communication to the reward. Regarding the amount of data sharing between the communicating agents $D_{shared}$, there are significant differences among the proposed study cases. Table \ref{table_evaluation_cases_metrics} shows that in study case 3 inter-agent communication does not provide any relevant value in aiding the exploration process and it achieves the lowest value among all the study cases. Conversely, the policies derived from the reward function proposed in study case 4 optimize inter-agent data transmission, as the exchanged maps most significantly enhance environmental knowledge acquisition. Study cases 1 and 2 yield similar values, which are lower compared to the optimal results from study case 4.

Eventually, Fig. \ref{fig:case_data} illustrates the results of the exploration performed by the study cases' policies in the same exploration arena. In study case 1 the agents tend to overlap many times focusing the exploration in the rightmost part of the map. On the contrary, case 4 is the one with less overlap between the agents' trajectories since the agents explore different regions of the map, while cases 2 and 3 tend to have lots of overlays between agents. Concerning the balance in the division of exploration work by agents, it can be stated that in this environment case 1 is more balanced since in the other three cases agent 2 does not contribute as the other agents to the exploration. On the right, some plots illustrate the information shared during network-based communication. Case 4 was the one that counted the most communication events in this environment. Moreover, as shown by the collected images, the communication networks have been very heterogeneous since it can be noted that the following networks have been formed: agent 1 and agent 2, agent 1 and agent 4, agent 3 and agent 4, agent 2 and agent 3, agent 1 and agent 3. The other study cases do not show satisfactory results concerning communication, except for case 2 in which a relevant communication pattern happens between agent 1 and agent 2.

Overall, considering the evaluated metrics and the simulation results depicted in Fig. \ref{fig:case_data}, it can be stated that study case 4 minimizes the overlap between the agents during the exploration task, reduces the number of steps to reach at least 90\% of exploration and proves to be robust. Moreover, Fig. \ref{fig:case_data} shows that it involves many different combinations of agents in the communication networks. 
\subsection{Simulation Results in Gazebo}
\label{sct:simulation_of_the_multi_agent_exploration_architecture_in_gazebo}
\label{sct:simulation_of_the_multi_agent_exploration_architecture_in_gazebo}
\begin{figure}[t]
    \centering
    \includegraphics[width=0.5\linewidth]{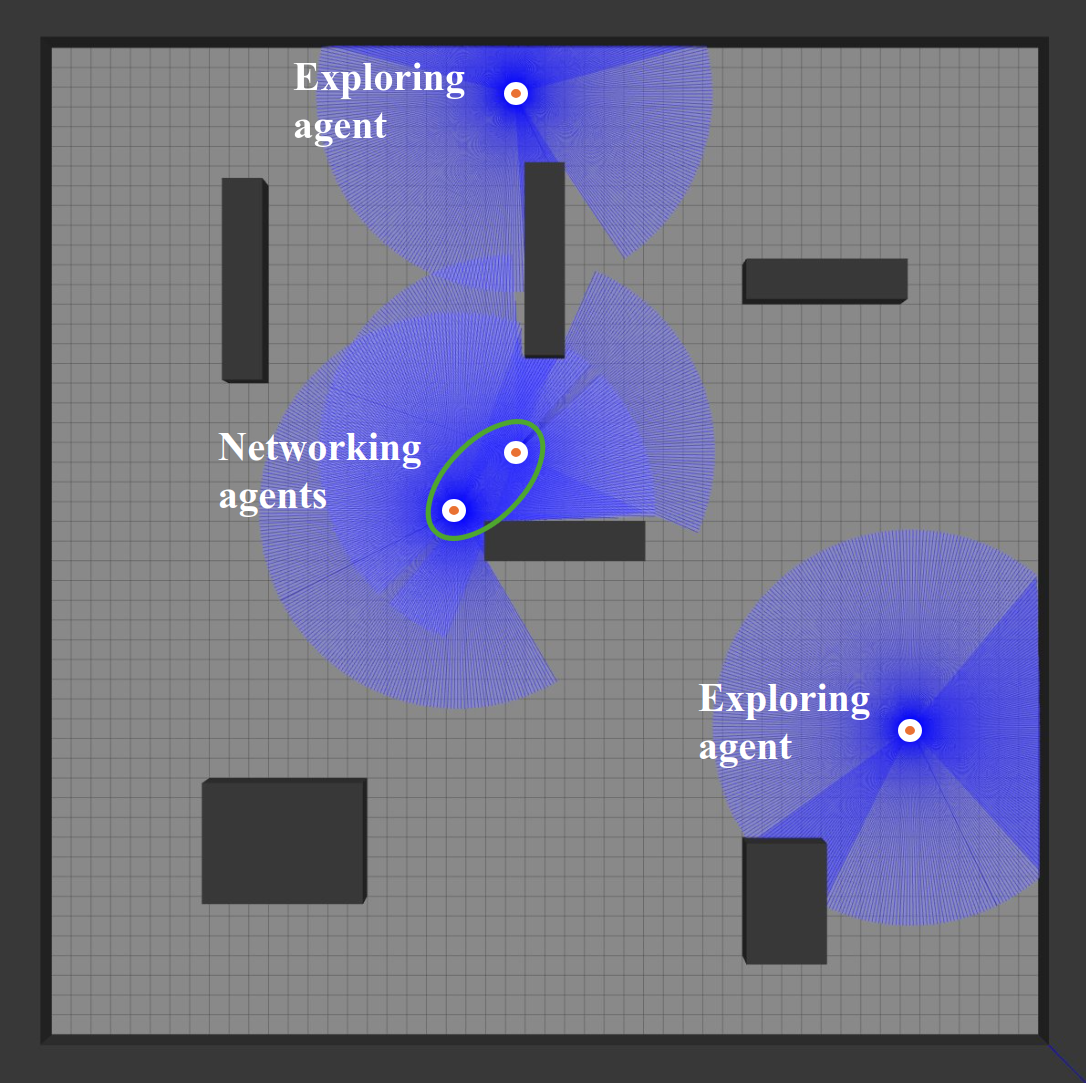}
    \caption{Exploration arena with obstacles (dark gray) and four deployed TurtleBot3 Burgers (white). Moreover, the laser scanner range is depicted in blue, and the green-outlined ellipse contains the networking agents}
\label{fig:gazebo_world_with_turtlebots}
\end{figure}

The performances of the trained agents' policies, whose parameters have been optimized following the procedure illustrated in Section \ref{subsection_training_setting}, have been assessed in executing the exploration task in a simulated environment designed with the Robot Operating System (ROS) 2 Humble Hawksbill \cite{ros2_humble} and the Gazebo multi-robot simulator, version 11.10.2 \cite{gazebo_classic}. Specifically, the simulation consists of four TurtleBot3 Burger robots, two-wheeled differential drive type platforms, which autonomously explore a suitably-designed arena filled with obstacles, as shown in Fig. \ref{fig:gazebo_world_with_turtlebots}, by executing the actions selected by the RL-based policies. This assessment aims to validate the framework's effectiveness in handling occupancy grids derived from the robot's laser scanner and the Simultaneous Localization and Mapping (SLAM) techniques. It also seeks to verify the mapping between each agent's actions and the robotic platform's movements, as well as the synchronization management among multi-agent systems. Indeed, all these aspects are not a concern in the implementation in Gymnasium \cite{openai_gym}. This section develops as follows: \ref{sbs:description_of_the_simulated_environment} briefly describes the exploration arena in Gazebo and the simulated robotic platforms, then \ref{sbs:ros2_architecture} investigates the details of the multi-agent architecture implementation using the Robotic Operating System, and finally \ref{sbs:simulation_results_in_gazebo} shows the exploration results obtained by using the trained policy with the simulated robots. 

\subsubsection{Description of the simulated environment}
\label{sbs:description_of_the_simulated_environment}
The simulated exploration arena has been designed using the Gazebo simulator and is shown in Fig. \ref{fig:gazebo_world_with_turtlebots}. In particular, the environment confined by four walls that the robots can explore is contained in the first octant of the 3-dimensional space. The navigable area is a square with dimensions of \SI{25}{m} $\times$ \SI{25}{m} and a cell size of $l=$ \SI{0.5}{m}. To simulate obstacles, six cuboids of various sizes, orientations, and positions have been placed within the arena, covering non-traversable areas as highlighted in Fig. \ref{fig:gazebo_world_with_turtlebots}. The robotic platforms used in this simulation are TurtleBot3 Burgers, which are equipped with a laser scanner to detect obstacles within the exploration environment.

\begin{figure*}[t]
    \centering
    \includegraphics[width=\linewidth]{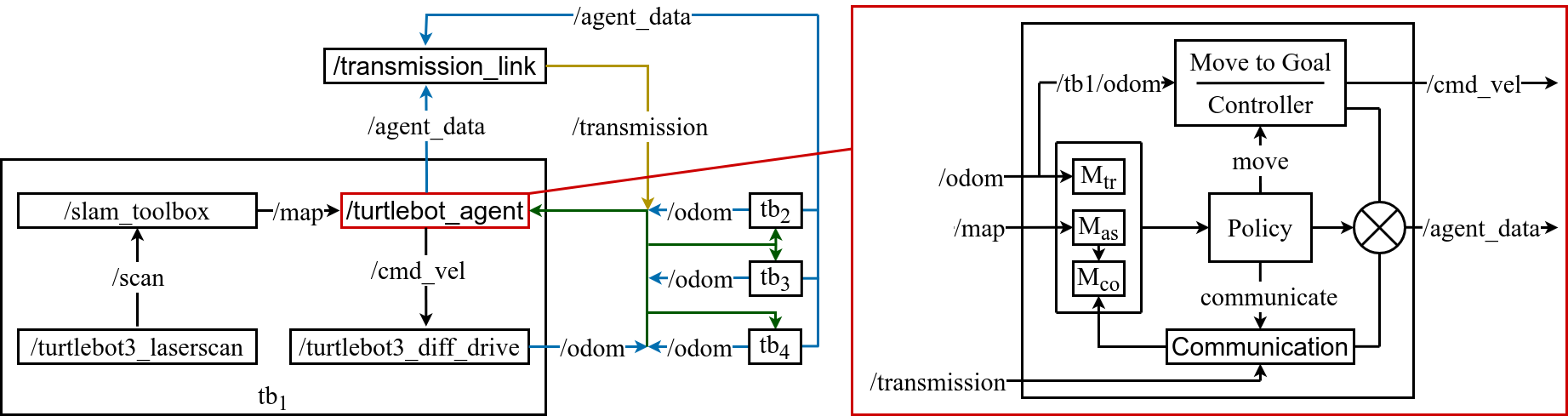}
    \caption{Overview of the main components of the ROS architecture for multi-agent exploration, focusing on the interactions between the robot $tb_1$'s nodes and the other agents through the $/transmission\_link$. The image uses color-coded links to differentiate the shared message types: blue for odometry data, yellow for messages on the $/transmission$ topic, and green links if both messages are shared. On the right is illustrated the structure of the $/turtlebot\_agent$ node, highlighting the information used to compute the agent's observations and actions.}
    \label{fig:ros2_architecture}
\end{figure*}

\subsubsection{ROS2 Architecture}
\label{sbs:ros2_architecture}
The Robot Operating System Humble Hawksbill has been used to define the architecture for the synchronization between the exploring and communicating agents, the mapping between the occupancy grid extracted through SLAM and the observation required by each agent's policy, and for implementing a controller that ensures the robot moves accordingly to the actions specified by the policy. Fig. \ref{fig:ros2_architecture} illustrates the most relevant entities in the ROS2 environment that constitute each agent. In particular, it shows the internal details of one TurtleBot, i.e., $tb_1$, and the node that establishes the transmission link between the networked agents. The following paragraphs outline the essential aspects of the ROS implementation for the interface with the RL-based policies. All these components have been implemented inside the \textit{agent\_data} node which is illustrated in Fig. \ref{fig:ros2_architecture}.\\
\textbf{\textit{Bridge between SLAM occupancy grids and the policies' observations}}: As per \ref{subsection_observation_space}, each agent's policy needs three maps to compute the next action, i.e., the agent-specific map, collaborative map, and transmission map. The occupancy grids necessary to obtain the first two maps have been generated by a SLAM node provided by the $\textit{slam\_toolbox}$ \cite{slam_toolbox_ros2} and associated with each robotic system. Then, the agent-specific and collaborative maps are computed extracting the states of the agent's nearby cells according to Algorithm \ref{alg:agent_specific_map_generation}.

\begin{algorithm}
\caption{Update of the Agent-Specific Map belonging to robot $i_k$}
\label{alg:agent_specific_map_generation}
\begin{algorithmic}[1]
\Require SLAM-based occupancy grid $\mathcal{M}_{i_k,slam}$, prior agent-specific map $\mathcal{M}_{i_k,as}$
\For{each $adjCell_{odom}$ in the agent's detection field}
    \State $adjCell_{map} \gets$ grid corresponding to $adjCell_{odom}$ in the $\mathcal{M}_{i_k,slam}$   
    \State $occProb \gets$ occupancy probability of $adjCell_{map}$
    \If{$-1 < occProb < 50$}
        \State $\mathcal{M}_{i_k,as}[adjCell_{odom}] \gets 0$ (free cell)
    \ElsIf{$occProb \geq 50$}
        \If{cell occupied by another agent}
            \State $\mathcal{M}_{i_k,as}[adjCell_{odom}] \gets 0$ (free cell)
        \Else
            \State $\mathcal{M}_{i_k,as}[adjCell_{odom}] \gets 1$ (obstacle)
        \EndIf
    \EndIf
\EndFor
\end{algorithmic}
\end{algorithm}

The agent's transmission map is created by identifying the positions of other agents within the agent's communication range.\\
\textit{\textbf{Interface between the agent's action and the robot's motion}}: The agent's action space comprises nine movement actions, specifying the neighboring cell to occupy in the next step in the execution of the reinforcement learning scheme. Therefore, a suitable interface has been defined to convert the policy's output into a desired configuration $q_d = [x_d, y_d]^T$, which is the center of the neighboring cell towards which the robot should navigate based on the movement chosen by the policy. The controller, formulated in Algorithm \ref{alg:control_strategy}, has been designed to set the robot velocity as a function of the actual configuration $q_t = [x_t, y_t, \theta_t]^T$ and the desired one $q_d$. The investigated control strategy for the navigation of the mobile robots assumes that the final robot's orientation is irrelevant. The relevant parameters used in Algorithm \ref{alg:control_strategy} are $\zeta=0.01$, $\gamma=0.1$, $K_v=1$ and $K_\omega=0.4$.\\
\begin{algorithm}
\caption{Control strategy for the navigation of a mobile robot to a desired configuration $q_d$}
\label{alg:control_strategy}
\begin{algorithmic}[1]
\Require desired agent's configuration $q_d = [x_d, y_d]^T$. distance threshold $\zeta$, orientation threshold $\gamma$, proportional gain for the linear velocity $K_v$, proportional gain for the angular velocity $K_{\omega}.$
\Ensure linear velocity $v$ and angular velocity $\omega$.
\For{$q_t = [x_t, y_t, \theta_t]^T$ obtained from agent's odometry}
    \State $e_x \gets x_d - x_t$
    \State $e_y \gets y_d - y_t$
    \State $e_\theta \gets \arctan(e_y, e_x) - \theta_t$  \Comment{Normalized in $[- \pi, \pi)$}
    \If{$\sqrt{e^{2}_{x} + e^{2}_{y}} \leq \zeta$}
        \State $v \gets 0$
        \State $\omega \gets 0$
    \ElsIf{$\lvert e_{\theta} \rvert > \gamma$}
        \State $v \gets 0$
        \State $\omega \gets K_{\omega} e_{\theta}$
    \Else
        \State $v \gets K_v \sqrt{e^{2}_{x} + e^{2}_{y}}$
        \State $\omega \gets 0$
    \EndIf
\EndFor
\end{algorithmic}
\end{algorithm}
%\begin{table}[b]
 %   \centering
  %  \caption{Parameters for the navigation controller}
    %\label{tab:controller_parameters}
    %\begin{tabular}{lcccc}
     %   \toprule
      %  Parameter & $\zeta$ & $\gamma$ & $K_v$ & $K_{\omega}$\\
       % \midrule
        %Value & 0.01 & 0.1 & 1 & 0.4\\
        %\bottomrule
    %\end{tabular}
%\end{table}
\begin{table*}
    \centering
    \caption{Starting positions of the simulated agents in the Gazebo-designed environment according to the different simulation setups}
    \label{tab:simulation_starting_positions}
    \begin{tabular}{lcccc}
        \toprule
        & $tb_1$ & $tb_2$ & $tb_3$ & $tb_4$ \\
        \midrule
        Setup 1 & [1.25, \, 1.25]   & [1.25, \, 23.75]  & [23.75, \, 1.25] & [23.75, \, 23.75] \\
        Setup 2 & [16.75, \, 6.25]  & [11.75, \, 23.25] & [11.75, \, 4.25] & [5.75, \, 2.25]   \\
        Setup 3 & [10.25, \, 20.75] & [6.75, \, 15.25]  & [4.25, \, 14.75] & [3.75, \, 2.25]   \\
        Setup 4 & [15.25, \, 23.25] & [0.75, \, 7.75]   & [17.75, \, 10.75] & [7.25, \, 21.75]  \\
        \bottomrule
    \end{tabular}
\end{table*}
\begin{algorithm*}
\caption{Execution Steps for the Simulation of the Proposed Multi-Agent Exploration Architecture in ROS2 and Gazebo}
\label{alg:implementation_turtlebots}
\begin{algorithmic}
\Require agent pool $\mathcal{I}$, robots $\{tb_k \mid i_k \in \mathcal{I}\}$, robots' deployment positions $\{p_0^{i_k} \mid i_k \in \mathcal{I}\}$, actors' networks $\{\pi^k(\theta^{i_k}) \mid i_k \in \mathcal{I}\}$, map discovery ratio $\rho \in \mathbb{N}$
\For{$i_k \in \mathcal{I}$}
\State Load on robot $tb_k$ the policy $\pi^k(\theta^{i_k})$ and initialize the agent-specific, collaborative, and transmission maps $\mathcal{M}_{i_k} = (\mathcal{M}_{i_k,as}, \mathcal{M}_{i_k,co}, \mathcal{M}_{i_k,tr})$
\EndFor
\State $j \gets 0$  
\State $\rho_j \gets 0$

\While{$\rho_j < \rho$ or agent collision with an obstacle}

    \State $j \gets j + 1$ 
    \For{$i_k \in \mathcal{I}$}
        \State Update the maps $\mathcal{M}_{i_k,as}$ and $\mathcal{M}_{i_k,co}$ using the occupancy grid computed by the robot $rb_k$'s SLAM node and considering only the regions around the robot within the detection range $r_d$
        \State Update the map $\mathcal{M}_{i_k,tr}$ using the positions of the robots $\{tb_w \mid i_w \in \mathcal{I} \setminus i_k\}$
        \State Compute the agent's action $a_j^{i_k}$ from the observation $\mathcal{M}_{i_k}$ by applying the policy $\pi^k(\theta^{i_k})$
        \If{$a_j^{i_k}$ involve a movement to a position $p_j^{i_k}$}
            \State Request the navigation controller to generate a velocity profile to be used by the robot $tb_k$ to navigate to the position $p_j^{i_k}$
        \ElsIf{$a_j^{i_k}$ implies the activation of the inter-agent communication}
            \State Send the map $\mathcal{M}_{i_k,co}$ to the other agents in the same communication network and collect their collaborative maps in $\mathcal{M}_{sh} = \{\mathcal{M}_{i_x,co} \mid i_x \in \mathcal{I} \iff a_j^{i_x} \text{ = communication}\}$
            \State Update the map $\mathcal{M}_{i_k,co}$ by merging it with the other agents' collected collaborative maps, namely $\mathcal{M}_{sh}$
        \EndIf 
    \EndFor
    \State $\rho_j \gets \max_{i_k \in \mathcal{I}} \left( \frac{\text{Number of discovered cells in } \mathcal{M}_{i_k,co}}{\text{Total number of grids}} \right)$
\EndWhile 
\end{algorithmic}
\end{algorithm*}
\textit{\textbf{Implementation of the inter-agent communication strategy}}:
The node $transmission\_link$ depicted in Fig. \ref{fig:ros2_architecture} synchronizes the execution of the reinforcement learning scheme among the robotic platforms exploring the simulated environment. It ensures that all agents wait until each other has executed their action and facilitates the exchange of collaborative maps between communicating agents within the same network. 

Finally, Algorithm \ref{alg:implementation_turtlebots} illustrates the overall scheme of working of the simulation with the four Turtlebot3 Burger robotic platforms in the Gazebo-based exploration arena.
\subsubsection{Simulation results}
\label{sbs:simulation_results_in_gazebo}
\begin{figure*}[]
    \centering
    \begin{subfigure}{0.48\linewidth}
        \centering
        \includegraphics[width=\linewidth]{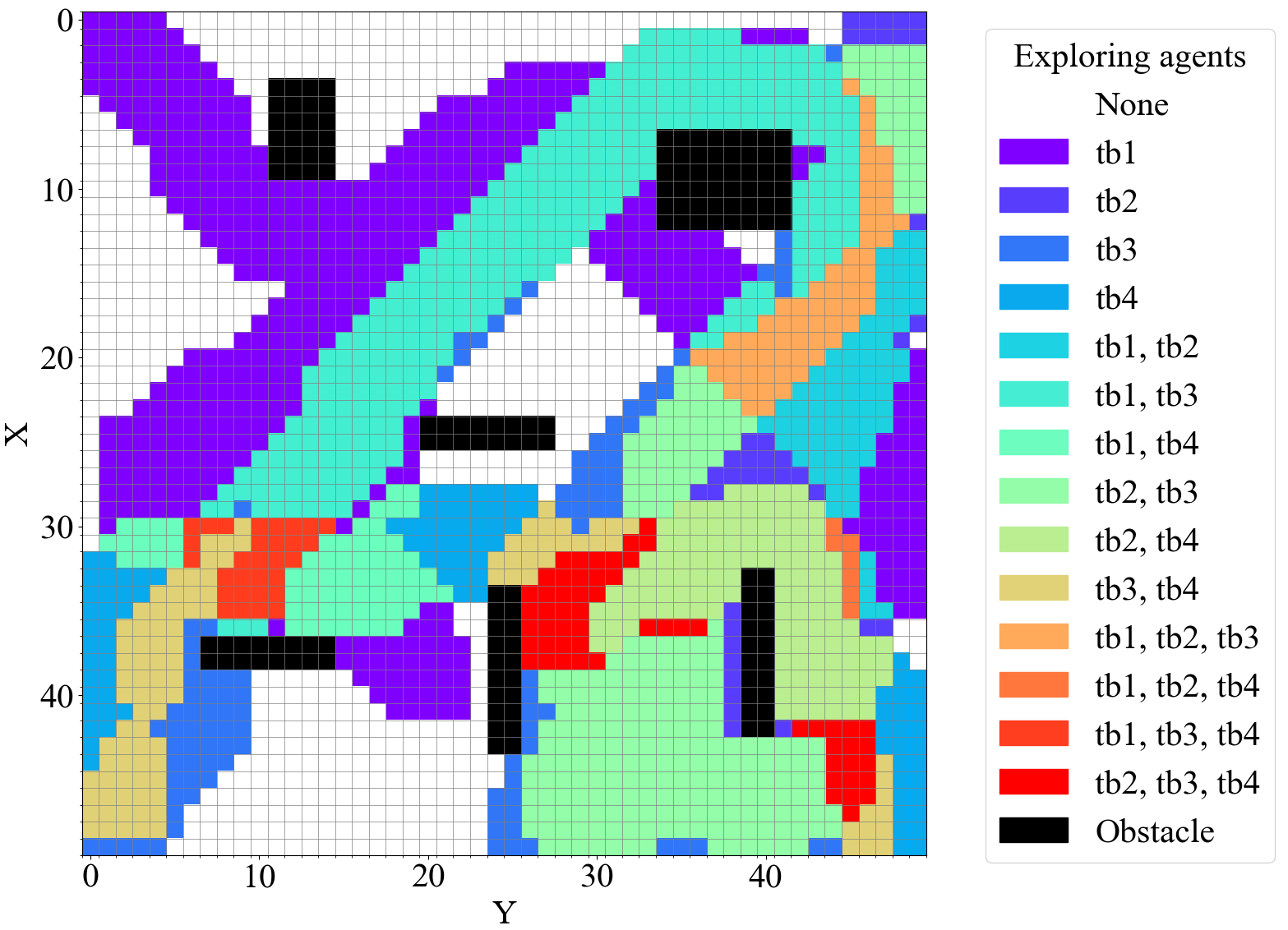}
        \caption{Regions explored by agents with the policies trained using the first study case}
        \label{fig:case1_gazebo}
    \end{subfigure}
    \hfill
    \begin{subfigure}{0.48\linewidth}
        \centering
        \includegraphics[width=\linewidth]{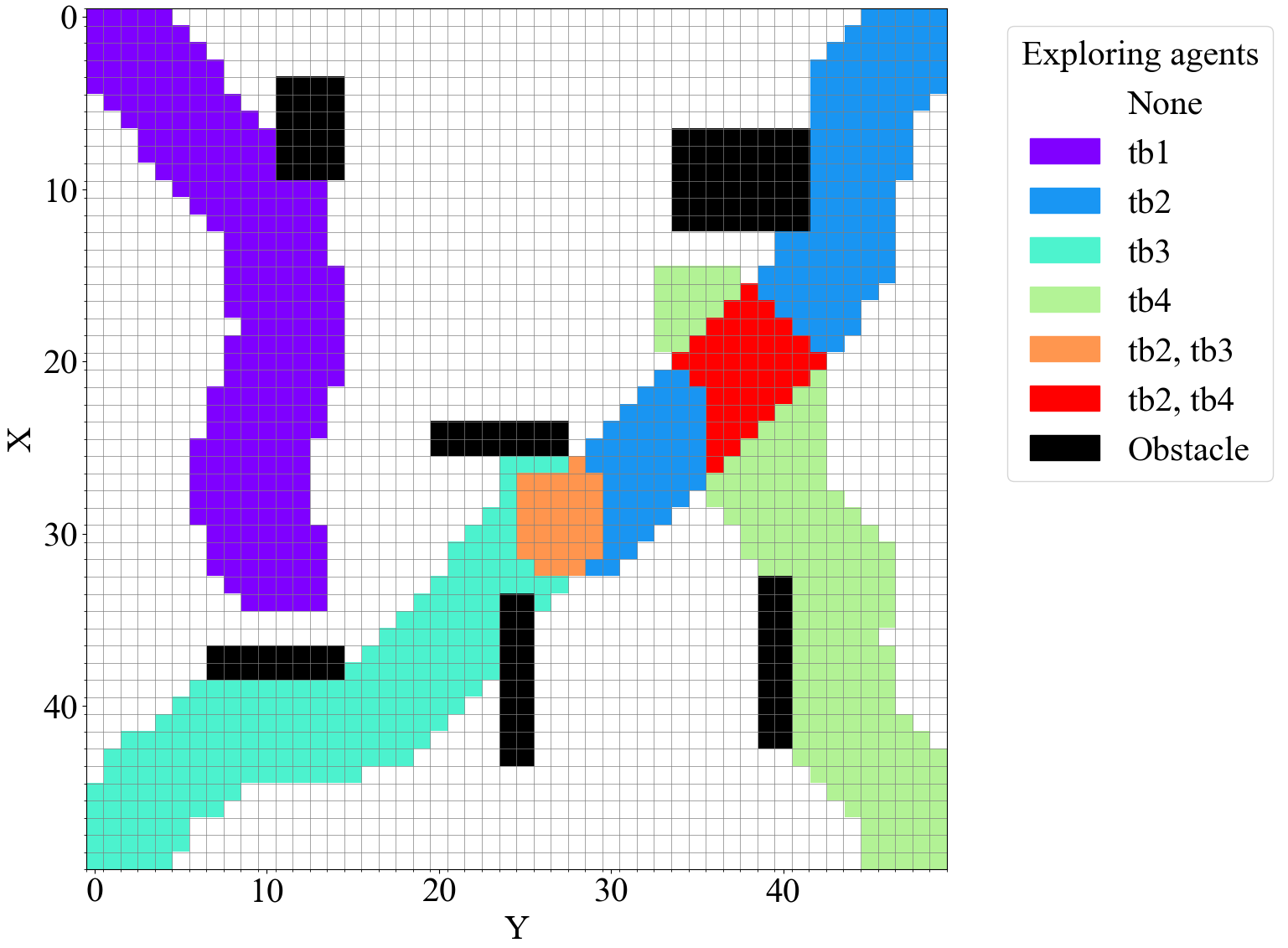}
        \caption{Regions explored by agents with the policies trained using the second study case}
        \label{fig:case2_gazebo}
    \end{subfigure}
    \begin{subfigure}{0.48\linewidth}
        \centering
        \includegraphics[width=\linewidth]{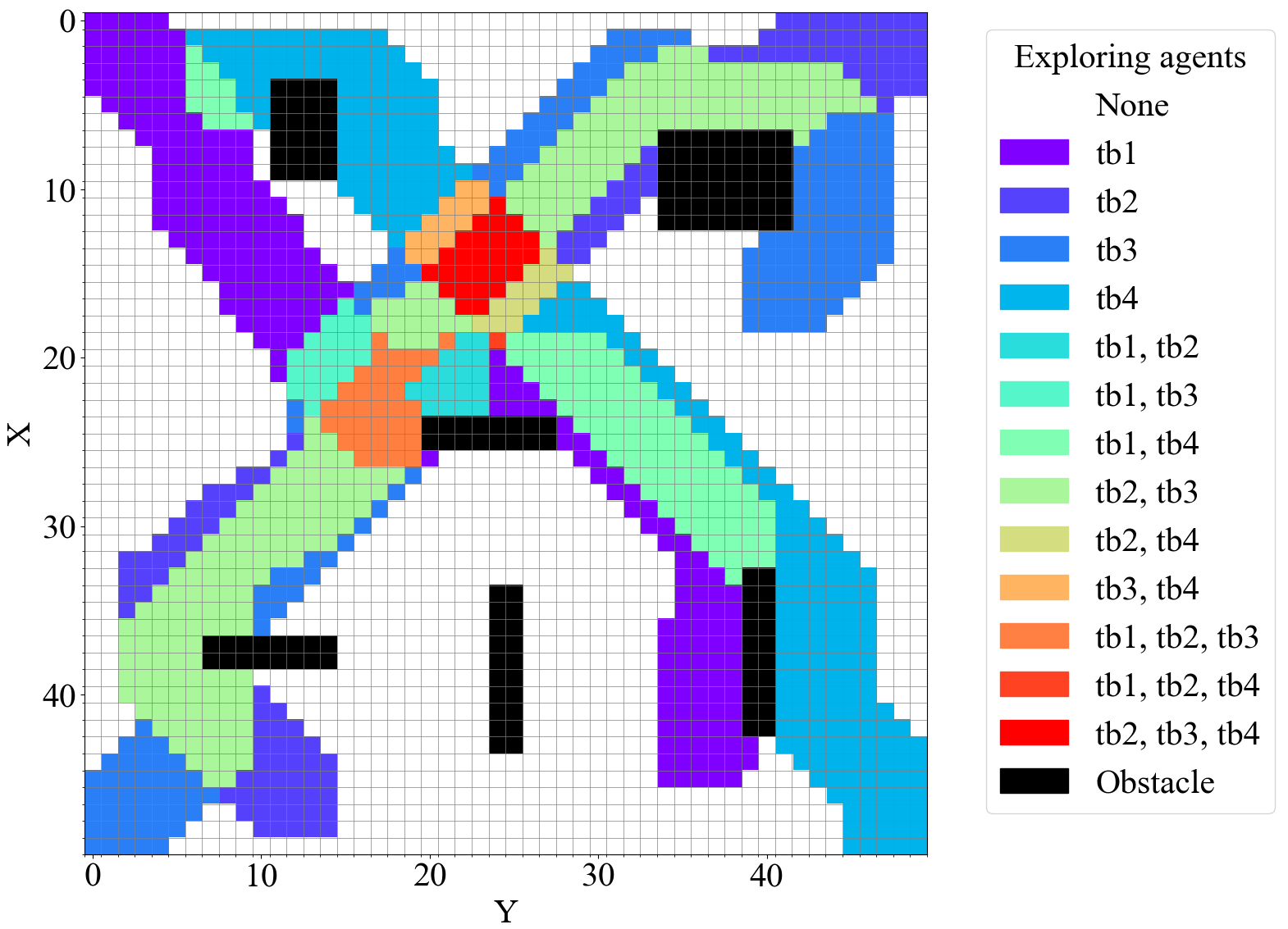}
        \caption{Regions explored by agents with the policies trained using the third study case}
        \label{fig:case3_gazebo}
    \end{subfigure}
    \hfill
    \begin{subfigure}{0.48\linewidth}
        \centering
        \includegraphics[width=\linewidth]{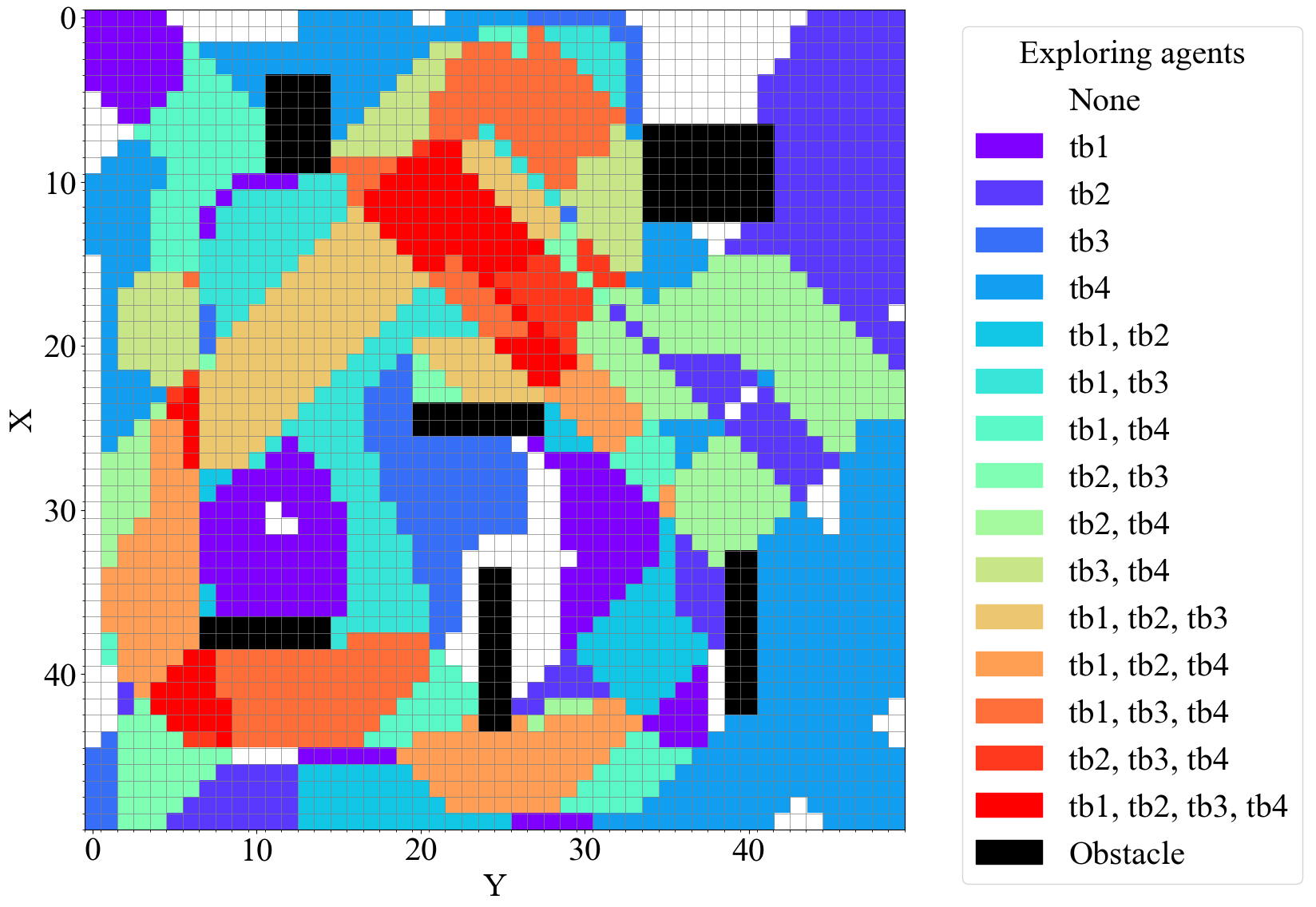}
        \caption{Regions explored by agents with the policies trained using the fourth study case}
        \label{fig:case4_gazebo}
    \end{subfigure}
    \caption{Maps of the Gazebo-designed environment depicting the regions explored by one or more agents during the simulation with starting agents' positions as per setup 1. The four illustrations show the cells that are occupied by obstacles (black), the grids that have not been explored by the agents (white), and the explored ones. These last elements are highlighted with different colors based on the clusters of agents that explored them, as indicated in the legends}
    \label{fig:overlapping_cases_gazebo}
\end{figure*}

The policies trained using the four study cases mentioned in \ref{subsection_synthesized_reward_functions} have been tested on four simulation setups in the Gazebo environment as depicted in Fig. \ref{fig:gazebo_world_with_turtlebots}. Each setup involves different starting positions of the agents, as indicated in Table \ref{tab:simulation_starting_positions}. The simulations were run until the agents successfully explored around $90\%$ of the navigable cells in the environment or at the first collision of a robotic platform with the obstacles arranged in the environment. Table \ref{table_evaluation_cases_metrics_gazebo} shows the map coverage and the overlap between agent-specific maps after each batch of simulations. It can be noticed that in all the setups the policies that were trained by rewarding inter-agent communication adequately, outperformed the first study case as shown by the map coverage ratio. However, the fourth study case, which is the best one according to the simulations in Gymnasium (\ref{sbs:simulation_results_in_gymnasium}), achieves better map coverage in two simulation setups out of the four proposed. Another measure that has been taken into consideration is the overlap between the agent-specific maps produced by each robotic platform as indicated both by the Jaccard coefficients in Table \ref{table_evaluation_cases_metrics_gazebo} and by the visual representation provided by Fig. \ref{fig:overlapping_cases_gazebo}. The former parameters tend to be low in all simulation setups and for all types of trained policies. However, although some high values are observed as map coverage increases, the Jaccard coefficients remain below 0.4. In particular, Fig. \ref{fig:overlapping_cases_gazebo} provides detailed views of the regions in the exploration arena that are inspected by multiple agents, focusing on all the study cases, and with agents' starting positions as per simulation setup 1. The figures indicate that policies trained with the fourth study case result in extensive exploration of large areas by one or a few agents, despite some central parts being examined by many robots, i.e., areas highlighted in orange and red. In contrast, policies trained with the first reward function lead to less overall surface exploration, with most regions being covered due to the overlapping efforts of different agents. Both study cases 2 and 3 generate maps that are not satisfactory since the majority of the free cells are not discovered as shown in Fig. \ref{fig:case2_gazebo} and \ref{fig:case3_gazebo}. Therefore, the simulation of the policies in the Gazebo environment shows that the communication terms in the reward functions may benefit the overall multi-agent exploration process. However, the overlapping between different agents is rather high.
% Table for Environment 1 and 2
\begin{table*}[]
    \centering
    \caption{Metrics collected from simulations based on different setups regarding agents' starting positions and different study cases (SC) used for policy training. In particular, this table illustrates the map coverage ratio, and the Jaccard coefficients $J_{i_k, i_j}$ between each pair of agent-specific maps $\mathcal{M}_{as,i_k}$ and $\mathcal{M}_{as,i_j}$ as described in \ref{sbs:simulation_results_in_gazebo}}
    \label{table_evaluation_cases_metrics_gazebo}
    
    \begin{tabular}{>{\centering\arraybackslash}p{1.3cm}*{8}{>{\centering\arraybackslash}p{1cm}}}
    \toprule
    & \multicolumn{4}{c}{Setup 1} & \multicolumn{4}{c}{Setup 2} \\
    \cmidrule(lr){2-5} \cmidrule(lr){6-9}
    & SC 1 & SC 2 & SC 3 & SC 4 & SC 1 & SC 2 & SC 3 & SC 4 \\
    \midrule
    Map coverage & $76.8\%$ & $37.7\%$ & $53.2\%$ & $91.4\%$ & $33.7\%$ & $34.4\%$ & $28.8\%$ & $52.2\%$ \\
    
    $J_{i_1, i_2}$ & 0.095 & 0.0 & 0.067 & 0.295 & 0.0 & 0.0 & 0.0 & 0.112 \\
    $J_{i_1, i_3}$ & 0.262 & 0.0 & 0.062 & 0.391 & 0.182 & 0.0 & 0.0 & 0.203 \\
    $J_{i_1, i_4}$ & 0.071 & 0.0 & 0.146 & 0.294 & 0.205 & 0.0 & 0.0 & 0.116 \\
    $J_{i_2, i_3}$ & 0.308 & 0.064 & 0.457 & 0.207 & 0.114 & 0.0 & 0.008 & 0.136 \\
    $J_{i_2, i_4}$ & 0.207 & 0.105 & 0.048 & 0.265 & 0.0 & 0.0 & 0.003 & 0.039 \\
    $J_{i_3, i_4}$ & 0.153 & 0.0 & 0.044 & 0.217 & 0.119 & 0.116 & 0.169 & 0.127 \\
    \bottomrule
\end{tabular}

    \begin{tabular}{>{\centering\arraybackslash}p{1.3cm}*{8}{>{\centering\arraybackslash}p{1cm}}}
        & \multicolumn{4}{c}{Setup 3} & \multicolumn{4}{c}{Setup 4} \\
        \cmidrule(lr){2-5} \cmidrule(lr){6-9}
        & SC 1 & SC 2 & SC 3 & SC 4 & SC 1 & SC 2 & SC 3 & SC 4 \\
        \midrule
        Map coverage & $64.6\%$ & $15.5\%$ & $69.8\%$ & $38.9\%$ & $63.3\%$ & $75.8\%$ & $39.2\%$ & $50.2\%$ \\
        $J_{i_1, i_2}$ & 0.137 & 0.0 & 0.093 & 0.035 & 0.119 & 0.126 & 0.0 & 0.066 \\
        $J_{i_1, i_3}$ & 0.110 & 0.0 & 0.093 & 0.0 & 0.080 & 0.175 & 0.0 & 0.0 \\
        $J_{i_1, i_4}$ & 0.115 & 0.0 & 0.102 & 0.0 & 0.102 & 0.087 & 0.194 & 0.0 \\
        $J_{i_2, i_3}$ & 0.117 & 0.327 & 0.043 & 0.038 & 0.056 & 0.139 & 0.010 & 0.201 \\
        $J_{i_2, i_4}$ & 0.377 & 0.0 & 0.101 & 0.06 & 0.135 & 0.161 & 0.0 & 0.0 \\
        $J_{i_3, i_4}$ & 0.100 & 0.0 & 0.069 & 0.311 & 0.335 & 0.058 & 0.145 & 0.101 \\
        \bottomrule
    \end{tabular}
\end{table*}
\begin{comment}
    Correctness                                     & $96.6\%$ & $98.7\%$ & $96.0\%$ & $97.2\%$ & $97.3\%$ & $96.9\%$ & $95.8\%$ & $97.5\%$ \\
    Correctness                                     & $97.4\%$ & $96.1\%$ & $96.3\%$ & $96.9\%$ & $97.3\%$ & $97.4\%$ & $96.0\%$ & $97.1\%$ \\
\end{comment}
\section{Conclusion}
This article presents a decentralized multi-agent reinforcement learning architecture for a group of homogeneous agents to navigate an unknown environment filled with obstacles. The predominant contribution of this work is the enhancement of the action space with inter-agent communication to train the robots to autonomously decide when to share data with the other agents and when to explore. To analyze the advantages of the proposed approach, we design different reward functions aimed at balancing knowledge-sharing among the agents and exploration behaviors. The results show that the exploration task performed by agents trained with reward functions implementing inter-agent communication reduces both the overlap in the explored areas and the time steps needed to explore the environment. Specifically, mitigation of task overlap is achieved by merging the knowledge accumulated by the different agents. Experiments with four Turtlebot3 Burgers in a Gazebo simulation environment confirmed the proposed strategy's effectiveness, though some critical challenges were observed compared to the exploration arena in Gymnasium. Future efforts will prioritize improving the multi-agent exploration scheme through real-world data, refining the policies, and leveraging heterogeneous agents.

\section{Statements and Declarations}
\paragraph{Funding} This work was partially supported by the Wallenberg AI, Autonomous Systems and Software Program (WASP) funded by the Knut and Alice Wallenberg Foundation, and by the European Union's Horizon Europe Research and Innovation Program, under the Grant Agreement No. 101119774 SPEAR.
\paragraph{Competing Interests} The authors declare that the authors have no competing interests as defined by Springer, or other interests that might be perceived to influence the results and/or discussion reported in this paper.
\paragraph{Author Contributions} Gabriele Calzolari, as the primary author, was responsible for the design and implementation of the methodologies, the execution of all simulations, and the preparation of the initial manuscript draft. Vidya Sumathy contributed to the conceptualization and formulation of the research framework, assisted in the design and execution of simulations, and participated in the manuscript preparation. Christoforos Kanellakis and George Nikolakopoulos supervised the overall research, offering critical guidance to both the primary and secondary authors. They also undertook the revision and editing of the manuscript. All authors reviewed and approved the final version of the paper.
\paragraph{Data Availability} This manuscript does not report data generation or analysis.

\bibliography{main}% common bib file

\end{document}